\documentclass[11pt]{article}
\usepackage{preprint}
\usepackage{times}
\usepackage{amsmath}
\usepackage{amssymb}
\usepackage{amsthm}
\usepackage{mathtools}
\usepackage{empheq}
\usepackage{graphicx}
\usepackage{booktabs}
\usepackage{makecell}
\usepackage{enumitem}
\usepackage{setspace}
\usepackage{algorithm}
\usepackage{algpseudocode}
\usepackage{natbib}
\usepackage{caption}
\usepackage{hyperref}
\hypersetup{
    colorlinks=true,
    linkcolor=blue,
    citecolor=blue,
    urlcolor=blue,
    filecolor=blue
}

\newcommand{\vect}[1]{\mathbf{#1}}
\newcommand{\syntha}{\textsc{synthetic-a}}
\newcommand{\synthb}{\textsc{synthetic-b}}
\newcommand{\credit}{\textsc{creditcard}}
\newcommand{\bank}{\textsc{bank}}
\newcommand{\bankfive}{\textsc{bank-5k}}
\newcommand{\bankforty}{\textsc{bank-40k}}
\newcommand{\adult}{\textsc{adult}}
\newcommand{\diabetes}{\textsc{diabetes}}
\newcommand{\swat}{\textsc{swat}}
\newcommand{\census}{\textsc{census1990}}

\begin{document}

    \onehalfspacing

    \title{Fast and effective algorithms for fair clustering at scale}

    \author{
      Claudio Mantuano, 
      Manuel Kammermann, 
      Philipp Baumann\textsuperscript{*} \\[6pt]
      Department of Business Administration, University of Bern,\\
      Engehaldenstrasse 4, 3012 Bern, Switzerland
    }
    \maketitle
    \renewcommand{\thefootnote}{}
    \footnotetext{\textsuperscript{*}Corresponding author. Emails: claudio.mantuano@unibe.ch, manuel.kammermann@outlook.com, philipp.baumann@unibe.ch.}
    \renewcommand{\thefootnote}{\arabic{footnote}}

    \begin{abstract}
        Clustering is an unsupervised machine learning task that consists of identifying groups of similar objects. It has numerous applications and is increasingly used in fairness-sensitive domains where objects represent individuals, such as customers, employees, or students. We address a fair clustering problem in which objects belong to protected groups. The problem consists of partitioning the objects into a predefined number of clusters while attaining a user-defined target level of fairness, meaning that each protected group is sufficiently represented in each cluster. The objective is to minimize the clustering cost, defined as the sum of squared Euclidean distances between the objects and the centers of their clusters. Since clustering cost and fairness are generally in conflict, managing the trade-off between them is essential in practical applications. Existing methods provide limited control over this trade-off and either fail to scale to large datasets or, when they scale, produce low-quality solutions. We propose a general framework for fair clustering that provides precise control over the cost-fairness trade-off and introduce three heuristics based on it. The first heuristic focuses on solution quality and the flexibility to incorporate additional constraints, the second improves scalability while retaining high solution quality, and the third is designed for maximum scalability, producing solutions for instances with millions of objects in seconds. The proposed heuristics outperform existing approaches in comprehensive numerical experiments on benchmark datasets. The source code of our heuristics and instructions for reproducing the experiments are publicly available on GitHub\footnote{\label{github:code}\url{https://github.com/claudio-mantuano/fair-clustering-at-scale}}.

    \end{abstract}

    \section{Introduction}\label{sec:introduction}
Clustering is an unsupervised machine learning task that consists of partitioning a dataset into subsets, called clusters, of similar objects. Clustering has a wide range of applications and is increasingly used to support or automate decisions that can have a direct impact on people’s lives. Common examples include bank loan disbursement \citep{tsai2010credit, bakoben2020identification, caruso2021cluster, baser2023credit}, human resources performance management \citep{wang2022optimization}, recruitment \citep{sivaram2010applicability, verma2017cluster}, student evaluation \citep{le2023review}, and college admissions \citep{mohamed2022clustering, cahapin2023clustering}. In such settings, vanilla clustering algorithms (i.e., conventional algorithms that do not explicitly account for fairness) may produce clusters in which certain groups of individuals are under-represented relative to their proportion in the dataset. This under-representation may introduce social disparities or amplify existing ones. According to the disparate impact doctrine of \cite{feldman2015certifying}, the outcomes of algorithmic decision-making systems should not adversely affect specific groups of individuals. Simply removing the sensitive feature (e.g., race) that identifies the protected groups (e.g., African-American, Asian, Caucasian) is generally ineffective. Other non-sensitive features (such as income or height) may still be correlated with the sensitive feature. For instance, \cite{lee2024algorithms} found that removing the sensitive feature ``race'' from a decision support system used for college admissions reduced the diversity of the top-ranked pool without improving the academic merit of the students in that pool. Building on the disparate impact doctrine, \cite{chierichetti2017fair} introduced the concept of fair clustering, marking the foundation of a now well-established line of research aimed at ensuring a fair assignment of protected groups to clusters. More recent studies have broadened the original scope of fair clustering by exploring its application to new domains, such as the security of cyber-physical systems \citep{sahoo2022evaluation, handa2022evaluating, obuzor2022scalable, chikhalia2022security}.

Several variants of fair clustering problems have been introduced in the literature (see \citealt{chhabra2021overview}). These variants differ mainly in how clustering cost is defined and how fairness is measured. We consider here the most common definition of the fair clustering problem, where center-based clustering cost (such as the well-known $k$-means clustering cost) and a group-level fairness notion based on \cite{chierichetti2017fair} (the balance) are employed. More specifically, the fair clustering problem considered here can be defined as follows. Given are a set of objects, represented by feature vectors, a set of sensitive features with two or more protected groups, and a number of clusters to be constructed. Each object belongs to one protected group of each sensitive feature. The goal is to partition the objects into clusters such that the $k$-means clustering cost, defined as the sum of squared Euclidean distances between the objects and the corresponding cluster centers, is minimized. The cluster centers are computed as the mean of the feature vectors of the objects assigned to each cluster. Fairness is measured for each cluster based on the distribution of protected groups within that cluster. A clustering solution is considered fair if each protected group is sufficiently represented in each cluster. The highest level of fairness is achieved when a protected group has the same representation in each cluster as it does in the entire dataset. The lowest level of fairness is achieved when a protected group is not represented at all in one of the clusters. We propose determining the required level of representation for each protected group using a single control parameter, which we refer to as the tolerance parameter. A tolerance value of zero requires the highest level of fairness, while a tolerance value of one imposes no representation requirement. Hence, the tolerance parameter allows us to control the cost-fairness trade-off. The fair clustering problem considered here is NP-hard, as it generalizes the $k$-means clustering problem (see \citealt{aloise2009np}). 

\cite{chierichetti2017fair} were the first to introduce fair clustering. They proposed a two-step approach that first constructs small fairness-preserving micro-clusters (fairlets) and then applies a vanilla clustering algorithm to group these fairlets. Following the seminal work of \cite{chierichetti2017fair}, a wide range of fair clustering methods has been developed. The development of algorithms has focused on three main research directions. The first research direction has focused on developing algorithms for extended or modified versions of the fair clustering problem (e.g., see \citealt{bera2019fair, schmidt2019fair, bercea2018cost}). The second research direction has focused on the development of algorithms for analyzing the trade-off between clustering cost and fairness (e.g., see \citealt{ziko2021variational, li2024optimization}). The third research direction has focused on developing scalable fair clustering approaches. \cite{backurs2019scalable} proposed a more efficient technique for constructing fairlets, which enables the fair clustering approach of \cite{chierichetti2017fair} to scale to much larger datasets. \cite{schmidt2019fair} introduced a preprocessing technique for constructing so-called coresets for fair clustering. The coresets are summaries of objects of the dataset and allow fair clustering to be sped up by reducing the input size. In contrast to fairlets, coresets are not fairness-preserving, which implies that fair clustering algorithms need to be applied instead of vanilla clustering algorithms. Despite the vast research effort, the existing algorithms still have limitations that hinder their use in practice. A limitation that affects most existing algorithms is that they are only applicable to specific variants of the fair clustering problem and lack the flexibility to incorporate practical constraints such as cardinality constraints or must-link and cannot-link constraints. A limitation of existing algorithms for analyzing the cost-fairness trade-off is that they rely on a weight parameter to combine the two criteria in the objective function. This approach offers only coarse and indirect control, making it difficult to attain specific fairness levels. A limitation of scalable algorithms is that they rely on approximating the original dataset using either fairlets or coresets, which can lead to information loss and, consequently, to a degradation in the quality of the resulting clustering solution.

In this article, we address the limitations discussed above by introducing a framework for fair clustering and three heuristics based on it. The framework is based on the well-known $k$-means decomposition scheme (see \citealt{lloyd1982least}), which alternates between an object assignment step and a cluster center update step until a stopping criterion is met. The three heuristics differ with respect to how the two steps are implemented, and therefore provide different advantages. The first heuristic is called mathematical programming-based fair clustering (MPFC) and focuses on solution quality and flexibility. In MPFC, the assignment step is formulated as a binary linear program (BLP) that includes fairness constraints that ensure the desired level of fairness in the assignment. The number of fairness constraints depends on the number of sensitive features and their protected groups. The constraints are designed so that the trade-off between fairness and clustering cost can be controlled by a single tolerance parameter. While solving the full clustering problem to optimality would be intractable even for medium-sized instances \citep{sauglam2006mixed, fleszar2008effective, zhou2019heuristic, amorosi2026mathematical}, restricting the use of integer programming to the assignment step allows the method to scale to instances with up to 100{,}000 objects. For the special case of a single sensitive feature, we propose a second heuristic that is substantially faster than MPFC while providing similar solution quality. The speedup is achieved by decomposing the assignment step into multiple subproblems, each of which can be solved efficiently using a minimum-cost flow algorithm. We refer to this approach as multi-stage minimum-cost flow-based fair clustering (MS-FlowFC). For some applications of fair clustering, million-scale datasets need to be processed (e.g., see \citealt{sahoo2022evaluation}). For such large-scale instances, we propose a third heuristic called scalable mathematical programming-based fair clustering (S-MPFC) that is even more scalable. S-MPFC combines an adapted version of MPFC with a preprocessing technique. The preprocessing technique aggregates nearby objects into representatives, which form a reduced dataset that is then clustered using an adapted version of MPFC. Our heuristics add to the literature in the following ways:

\begin{itemize}[noitemsep, topsep=0pt]
    \item We propose highly flexible approaches that can be readily adapted to application-specific requirements. First, all methods are based on the $k$-means decomposition scheme, which naturally extends to alternative center-based objectives such as $k$-median and $k$-medoids. Second, the assignment step in the BLP-based approaches (MPFC and S-MPFC) can be easily extended to incorporate additional constraints, including cardinality constraints as well as must-link and cannot-link constraints.
    \item We propose approaches that offer precise control over the desired level of fairness by adjusting the constraints of the assignment step using a tolerance parameter. This overcomes the limitations of existing approaches that rely on a weight parameter in the objective function to trade off fairness against clustering cost.  
    \item We propose novel techniques to achieve scalability in fair clustering, namely a decomposition strategy for the assignment step that does not rely on an integer programming solver and a preprocessing technique to aggregate the input data. The preprocessing technique is similar to existing ones in that it approximates the dataset using small subsets of objects. However, it differs in that it does not impose any restrictions on the composition of the subsets, that is, they do not need to be fair and may contain objects from different protected groups.   
    \item We additionally provide two exact approaches for benchmarking: a mixed-integer quadratically constrained programming (MIQCP) model solved with Gurobi Optimizer and a set-variable-based (SetVars) formulation solved with Hexaly. To the best of our knowledge, these are the first exact methods for the fair $k$-means clustering problem considered here. 
\end{itemize}

In a comprehensive computational analysis, we assess the performance of the proposed approaches MPFC, MS-FlowFC, and S-MPFC. The computational analysis is based on two synthetically generated and six real-world datasets from the literature and includes as benchmark methods the scalable fair clustering algorithm of \cite{backurs2019scalable} (SFC) and the order-and-cut algorithm of \cite{li2024optimization} (O\&C). In addition to the two methods from the literature, we further include as baselines the two exact fair clustering approaches proposed in this paper, as well as Lloyd’s algorithm for vanilla clustering. The results demonstrate that the proposed approaches can find optimal solutions on small problem instances and consistently, often substantially, outperform the two methods from the literature in both solution quality and running time. For example, on the largest dataset, containing nearly 2.5 million objects, the S-MPFC approach achieves higher-quality solutions than the method of \cite{backurs2019scalable}, while reducing running time by 99.70\%. In addition, the results demonstrate that the trade-off between fairness and clustering cost can be controlled more effectively than with the benchmark approach of \cite{li2024optimization}.  

The remainder of this article is structured as follows. Section~\ref{sec:fair_clustering} describes the fair clustering problem, introducing the notation and providing an illustrative example. Section~\ref{sec:literature} provides an overview of the literature on fair clustering, while Section~\ref{sec:proposed_approaches} presents the proposed heuristic approaches. Section~\ref{sec:numerical_experiments} details the numerical experiments conducted to evaluate the proposed methods and discusses the numerical results. Finally, Section~\ref{sec:conclusions} presents our conclusions and directions for future research.

\section{Fair clustering}\label{sec:fair_clustering}
Section~\ref{subsec:problem_description} describes the fair clustering problem and Section~\ref{subsec:illustrative_example} provides an illustrative example.

\subsection{Problem description}\label{subsec:problem_description}

Given is a dataset $X = \{\vect{x}_1, \vect{x}_2, \dots, \vect{x}_n\}$ consisting of $n$ objects, where each object $\vect{x}_i \in \mathbb{R}^d$ is described by $d$ non-sensitive features. In addition, one or multiple sensitive features (e.g., race) are given. We denote the set of sensitive features by $S$. Each sensitive feature $s \in S$ has two or more protected groups $g \in \mathcal{G}_s$ (e.g., African-American, Asian, Caucasian), and each object belongs to exactly one protected group of each sensitive feature. The goal is to partition the objects into $k$ clusters such that the fairness constraints related to the protected groups are satisfied and the $k$-means clustering cost is minimized. The $k$-means clustering cost is defined as the sum of squared Euclidean distances between each object and the center of its assigned cluster. The center of a cluster $C_j$, for $j=1,\ldots,k$, is computed as the mean of the non-sensitive feature values of all objects assigned to cluster $C_j$. Next, we describe the fairness constraints imposed for each sensitive feature $s \in S$. 

The fairness level of a clustering solution with respect to a sensitive feature $s \in S$ depends on the level of representation achieved by each protected group $g \in \mathcal{G}_s$ across all clusters. The level of representation of a protected group is measured using the notion of balance, originally introduced by \cite{chierichetti2017fair} for the case of a single sensitive feature with two protected groups. A higher balance indicates a higher level of representation and, consequently, a higher level of fairness. Since we consider multiple sensitive features with two or more protected groups, we extend the definition of balance accordingly. Given a clustering solution $\mathcal{C}$, its balance with respect to the sensitive feature $s \in S$ is defined as the minimum balance among all clusters $C_j \in \mathcal{C}$, with $j=1,\dots,k$:
\begin{equation}
	\label{eq:clustering_balance}
	B_s(\mathcal{C})=\min_{j =1,\dots,k} B_s(C_j),
\end{equation}

where $B_s(C_j)$ is the balance of cluster $C_j$. To compute $B_s(C_j)$, we consider all pairs of distinct protected groups $g, g' \in \mathcal{G}_s$, with $g \ne g'$, of the sensitive feature $s$. For each pair, we compute the ratio between the number of cluster members belonging to protected group $g$ and the number of cluster members belonging to protected group $g^{\prime}$, and vice versa. The balance $B_s(C_j)$ is then defined as the minimum over all computed ratios. Let $G_{gs}$ denote the set of objects belonging to protected group $g$ of sensitive feature $s$. Then, $B_s(C_j)$ is computed as follows:
\begin{equation}
	\label{eq:cluster_balance}
	B_s(C_j)=\min_{\substack{g, g' \in \mathcal{G}_s \\ g \ne g'}} \left(\frac{|C_j \cap G_{gs}|}{|C_j \cap G_{g^{\prime}s}|}\right).
\end{equation}

A clustering solution $\mathcal{C}$ is considered fair if, for each sensitive feature $s \in S$, the balance $B_s(\mathcal{C})$ is equal to or greater than a predefined target balance. We propose determining the target balance for each sensitive feature $s \in S$ based on a single parameter $\lambda \in [0, 1]$, which we refer to as the tolerance parameter. Smaller values of $\lambda$ lead to higher target balances for all sensitive features and, consequently, to a higher level of fairness. The target balance $\overline{B}_s(X, \lambda)$ for sensitive feature $s \in S$ of dataset $X$, given $\lambda$, is defined as follows:  
\begin{equation}
	\label{eq:target_balance}
	\overline{B}_s(X, \lambda) = (1-\lambda) B_s(X),
\end{equation}

where $B_s(X)$ denotes the dataset balance of sensitive feature $s$, which is computed as follows:
\begin{equation}
	\label{eq:dataset_balance}
	B_s(X)=\min_{\substack{g, g' \in \mathcal{G}_s \\ g \ne g'}} \left(\frac{|G_{gs}|}{|G_{g^{\prime}s}|}\right).
\end{equation}

If $\lambda=0$, the resulting target balance for sensitive feature $s$ is equal to the dataset balance for sensitive feature $s$. If $\lambda=1$, the resulting target balance for each sensitive feature $s$ is equal to zero, and the problem reduces to a vanilla clustering problem. This approach allows controlling the level of fairness of the clustering solution across all sensitive features by varying a single parameter $\lambda$. Note that, depending on the number of clusters $k$, it may not be always possible to achieve the dataset balance for each sensitive feature. The dataset balance is therefore only an upper bound on the maximum balance that can be achieved. Hence, for some (small) values of $\lambda$, no feasible solution may exist for a given number of clusters $k$. In our experimental analysis, we therefore replace the dataset balance $B_s(X)$ with a value $B_s(X, k)$ that is close to the dataset balance but guarantees feasibility for $\lambda = 0$, given the number of clusters $k$ (see Eq.~\eqref{eq:maximum_balance}). 

\begin{figure}[t]
    \centering
    \includegraphics[width=1\textwidth]{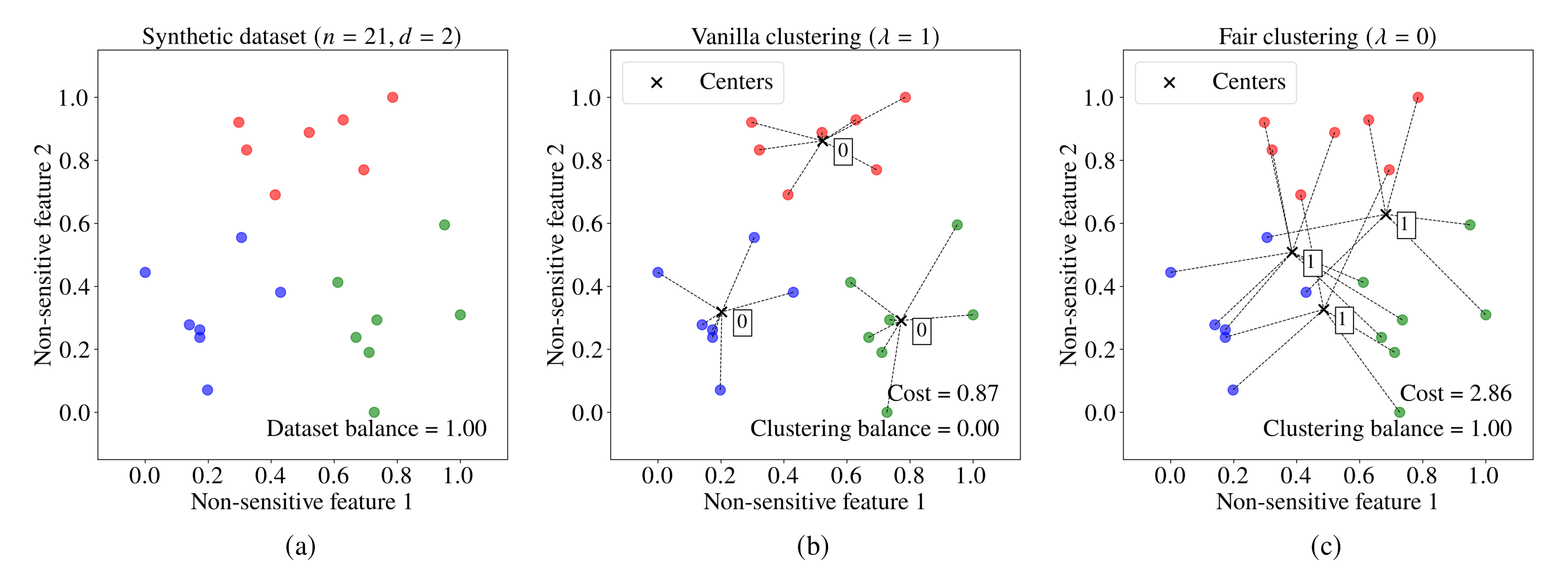}
    \caption{(a) synthetic dataset; (b) vanilla clustering optimal solution; (c) fair clustering optimal solution.}
    \label{fig:illustrative_example}
\end{figure}

\subsection{Illustrative example}\label{subsec:illustrative_example}

We present an example to illustrate the notation and definitions introduced in Section~\ref{subsec:problem_description}. Fig.~\ref{fig:illustrative_example} (a) shows a synthetic dataset $X$ composed of $n=21$ objects, each described by $d=2$ non-sensitive features. Additionally, each object is associated with a sensitive feature $s$ that has three protected groups. We use the colors green, blue, and red to indicate these protected groups. The protected groups are of equal size, which yields a dataset balance $B_s(X)=1$. The number of clusters is set to $k=3$. In this example, the dataset balance $B_s(X)=1$ can be achieved. According to Eq.~\eqref{eq:target_balance}, given the tolerance parameter $\lambda$, the target balance is $\overline{B}_s(X, \lambda)=0$ for the vanilla clustering problem ($\lambda=1$) and $\overline{B}_s(X, \lambda)=1$ for the fair clustering problem ($\lambda=0$). 

Fig.~\ref{fig:illustrative_example} (b) and Fig.~\ref{fig:illustrative_example} (c) illustrate the optimal solutions for vanilla clustering and fair clustering, respectively. We obtained these optimal solutions by solving the MIQCP formulation described in \ref{app:exact}. Black crosses denote cluster centers, while dashed lines indicate the assignment of each object to its respective center. The labels next to the centers report the balance of each cluster, computed as in Eq.~\eqref{eq:cluster_balance}. The clustering cost and balance of the solutions are shown in the bottom-right corner of each figure. In this example, vanilla clustering produced an unfair solution with clustering balance equal to zero, as each cluster contained objects from only a single protected group. In contrast, fair clustering achieved the fairest solution, albeit at a higher cost.

\section{Literature review}\label{sec:literature}

\cite{chierichetti2017fair} introduced fair clustering. They proposed balance as a measure of fairness, because it is consistent with the disparate impact doctrine of \cite{feldman2015certifying}. \cite{chierichetti2017fair} developed a two-step approach for fair $k$-center and $k$-median clustering. In the first step, which is referred to as fairlet decomposition, the dataset is partitioned into small subsets (fairlets) that satisfy fairness. In the second step, a vanilla clustering algorithm is applied to cluster these fairlets. This two-step approach is limited to a single sensitive feature with two protected groups. Building on this work, the literature on fair clustering has expanded rapidly and can be broadly categorized into three main directions. For a comprehensive overview, we refer the reader to \cite{chhabra2021overview}. 

The first direction of research has focused on introducing new variants of fair clustering problems. These problems are either extensions of the fair clustering problem introduced by \cite{chierichetti2017fair} or new types of fair clustering problems. The proposed problems introduce novel notions of fairness \citep{ahmadian2019clustering, abraham2019fairness, kleindessner2019fair, esmaeili2021fair, lawless2024fair}, some of which generalize to multiple sensitive features and protected groups \citep{rosner2018privacy, ahmadian2019clustering, abraham2019fairness, bera2019fair}. Moreover, problems with alternative clustering objectives, such as $k$-means, $k$-medoids, and $k$-supplier \citep{bercea2018cost, bera2019fair, quy2021fair}, as well as problems with additional constraints such as capacity constraints on clusters \citep{quy2021fair, tran2023mpfcc} have been studied.  

The second direction of research has investigated strategies to trade off clustering cost and fairness. \cite{abraham2019fairness} proposed to combine the clustering cost and fairness as a weighted sum in the objective function. Several subsequent studies have also combined clustering cost and fairness in the objective function and used a weight parameter to control the trade-off \citep{ziko2021variational, simoes2024towards, li2024optimization}. Other studies explored different strategies. \cite{esmaeili2021fair} proposed an approach that optimizes the fairness objective subject to a constraint on the clustering cost. \cite{liu2022stochastic} proposed a modified version of the mini-batch $k$-means algorithm for fair clustering that approximates the Pareto front by varying a control parameter of the algorithm that determines how much effort is spent on optimizing clustering cost versus optimizing fairness. \citet{hakim2024fairness} proposed a dynamic programming approach that constructs the exact Pareto front between clustering cost and fairness for fixed cluster centers. Combined with a vanilla clustering algorithm for center selection, it yields an approximate Pareto front.

The third direction of research has focused on improving the scalability of fair clustering methods. Several works have built on the idea of \cite{chierichetti2017fair} to first construct subsets of similar objects in a preprocessing step and then cluster the subsets rather than the individual objects. \cite{backurs2019scalable} introduced a nearly-linear time algorithm to perform the fairlet decomposition proposed by \cite{chierichetti2017fair}. \cite{schmidt2019fair} introduced a procedure to construct different types of subsets of similar objects, which are referred to as coresets. In contrast to fairlets, which must be fairness-preserving, the objects in coresets only need to belong to the same protected group. However, clustering coresets requires a fair clustering algorithm, whereas fairlets can be clustered using vanilla clustering algorithms. The approach of \cite{schmidt2019fair} was extended by \cite{huang2019coresets} to handle multiple sensitive features and protected groups. \cite{bandyapadhyay2024coresets} proposed a technique to construct coresets whose size does not grow exponentially with the dimension of the dataset. Another strategy to tackle large-scale instances of fair clustering problems is to decompose the original problem into simpler subproblems. \cite{tran2023mpfcc} applied the $k$-means decomposition scheme to decompose a fair capacitated clustering problem into an object assignment subproblem and a cluster center update subproblem. In their approach, the object assignment subproblem is formulated as a mathematical programming problem. The $k$-means decomposition has also been used for other fair clustering problems \citep{simoes2024towards, lawless2024fair} or other constrained clustering problems \citep{baumann2019binary, baumann2020binary, baumann2025algorithm}. 

The present study builds on works from the first research direction by considering the fairness notion of balance in a general setting with multiple sensitive features and multiple protected groups. Moreover, the algorithms we propose are flexible to incorporate practical constraints and can consider alternative clustering objectives, such as $k$-median or $k$-medoids with minor adjustments. We contribute to the second research direction by introducing a framework that offers higher precision to control the trade-off between clustering cost and fairness compared to existing approaches. Finally, we contribute to the third research direction by introducing two novel techniques to scale fair clustering algorithms to large-scale datasets. The first technique is based on solving a sequence of minimum-cost flow problems and the second technique is based on building subsets of objects without imposing any restrictions on their composition. 

\section{Proposed approaches}\label{sec:proposed_approaches}

\begin{figure}[t]
    \centering
    \includegraphics[width=0.95\textwidth]{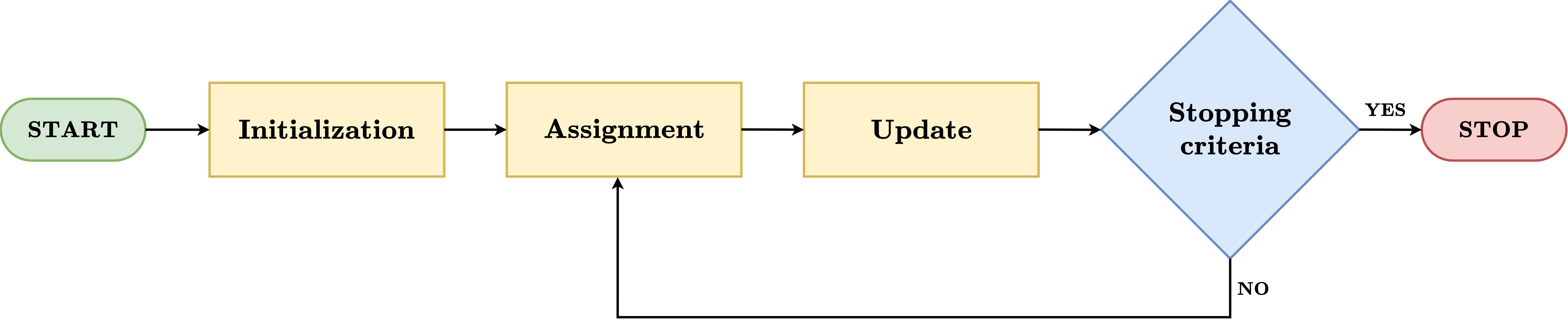}
    \caption{Decomposition scheme used by the proposed heuristics.}
    \label{fig:decomposition_scheme}
\end{figure}

In this section, we propose three center-based heuristics for the fair clustering problem. In center-based approaches, each cluster is represented by its center. All three heuristics follow the well-known decomposition scheme illustrated in Fig.~\ref{fig:decomposition_scheme}. In Section~\ref{subsec:decomposition_scheme}, we first describe the decomposition scheme. Then, we explain the three heuristics in Sections~\ref{subsec:mpfc}, \ref{subsec:ms_flowfc}, and \ref{subsec:s_mpfc}, respectively. 

\subsection{Decomposition scheme}\label{subsec:decomposition_scheme}

The idea of the decomposition scheme is to decouple the positioning of cluster centers from the assignment of objects to cluster centers. The heuristics start with the initialization of the cluster center positions. This initialization is done by selecting $k$ objects using the randomized $k$-means++ method of \cite{arthur2007k}. The positions of these $k$ selected objects serve as initial positions for the cluster centers. The heuristics then alternate between an assignment and an update step until a stopping criterion is met. In the assignment step, the objects are assigned to the cluster centers. In the update step, the positions of the centers are adjusted according to the results of the preceding assignment step. The difference between the three heuristics is that they use different approaches to perform the assignment and update steps. These approaches are described in Sections~\ref{subsec:mpfc}, \ref{subsec:ms_flowfc}, and \ref{subsec:s_mpfc}. In the update step, the position of each center is adjusted such that it is located at the center of gravity of its assigned objects. This adjustment optimizes the $k$-means objective. However, the framework readily extends to other clustering objectives by modifying only the update step. At each iteration $t$, the heuristics verify the following three stopping criteria: (i) the maximum running time \texttt{MaxTime} is reached, (ii) the solution improvement falls below the threshold $\Delta \geq 0$, or (iii) the maximum number of iterations \texttt{MaxIter} is attained. The improvement of the clustering cost at the end of iteration $t$, denoted by $\delta^{(t)}$, is computed as follows:
\begin{equation}
	\label{eq:improvement}
	\delta^{(t)} = 1 - \frac{\text{Cost}^{(t)}}{\text{Cost}^{(t-1)}},
\end{equation}

where $\text{Cost}^{(t)}$ is the clustering cost at the end of iteration $t$, and $\text{Cost}^{(t-1)}$ is the clustering cost at the end of iteration $t-1$. If at least one stopping criterion is met, the heuristics terminate. 

Since the $k$-means++ initialization is a randomized procedure, different solutions can be obtained by applying the heuristics with different random seeds. In our experiments presented in Section~\ref{sec:numerical_experiments}, we therefore apply our heuristics multiple times, each time with a different random seed. 

\subsection{Mathematical programming-based fair clustering (MPFC)}\label{subsec:mpfc}

In the first heuristic, referred to as mathematical programming-based fair clustering (MPFC), the assignment step is formulated as a binary linear programming (BLP) problem. In this BLP, the objects $\vect{x}_i$, for $i=1,\dots,n$, are assigned to cluster centers $\vect{\mu}_j$, for $j=1,\dots,k$. The positions of these cluster centers are fixed. The cost of assigning object $\vect{x}_i$ to center $\vect{\mu}_j$, denoted by $c_{ij}$, corresponds to the squared Euclidean distance between them. The binary decision variables $x_{ij}$ take the value one if object $i$ is assigned to center $j$, and zero otherwise. The goal is to obtain an assignment that satisfies the target balance while minimizing the sum of squared Euclidean distances between centers and their assigned objects. The following formulation (BLP) is solved in the assignment step of MPFC.
\newpage
\begin{empheq}[left=\text{(BLP)}\empheqlbrace]{align}
    \min \quad & \sum_{i=1}^n\sum_{j=1}^k  c_{ij} x_{ij} \label{mpfc_a}\\
    \text{s.t.} \quad & \sum_{j=1}^k x_{ij} = 1 && (i=1,\ldots,n) \label{mpfc_b}\\
    & \sum_{i=1}^n x_{ij} \geq 1 && (j=1,\ldots,k) \label{mpfc_c}\\
    & \sum_{i \in G_{gs}} x_{ij} \geq \overline{B}_s(X, \lambda) \sum_{i \in G_{g's}} x_{ij} && (g,g' \in \mathcal{G}_s: g \ne g';\ j=1,\ldots,k;\ s \in S) \label{mpfc_d}\\
    & x_{ij} \in \{0,1\} && (i=1,\ldots,n;\ j=1,\ldots,k) \label{mpfc_e}
\end{empheq}

The objective function (\ref{mpfc_a}) minimizes the total squared Euclidean distance between the centers and their assigned objects. Constraints (\ref{mpfc_b}) ensure that each object is assigned to exactly one cluster center, while constraints (\ref{mpfc_c}) guarantee that no cluster remains empty. Constraints (\ref{mpfc_d}) enforce fairness within each cluster by imposing a lower bound on the clustering balance for each sensitive feature. Specifically, they translate the definition of cluster balance in Eq.~\eqref{eq:cluster_balance} into a linear constraint: the ratio between the number of objects from any two protected groups of the same sensitive feature must be greater than or equal to the target balance of that sensitive feature ($\overline{B}_s(X, \lambda)$, see Eq.~\eqref{eq:target_balance}). This condition applies to every cluster, thereby ensuring that the clustering balance, computed as in Eq.~\eqref{eq:clustering_balance}, achieves the target balance. Finally, constraints (\ref{mpfc_e}) define the domain of the decision variables.

\begin{figure}[t]
    \centering
    \includegraphics[width=1\textwidth]{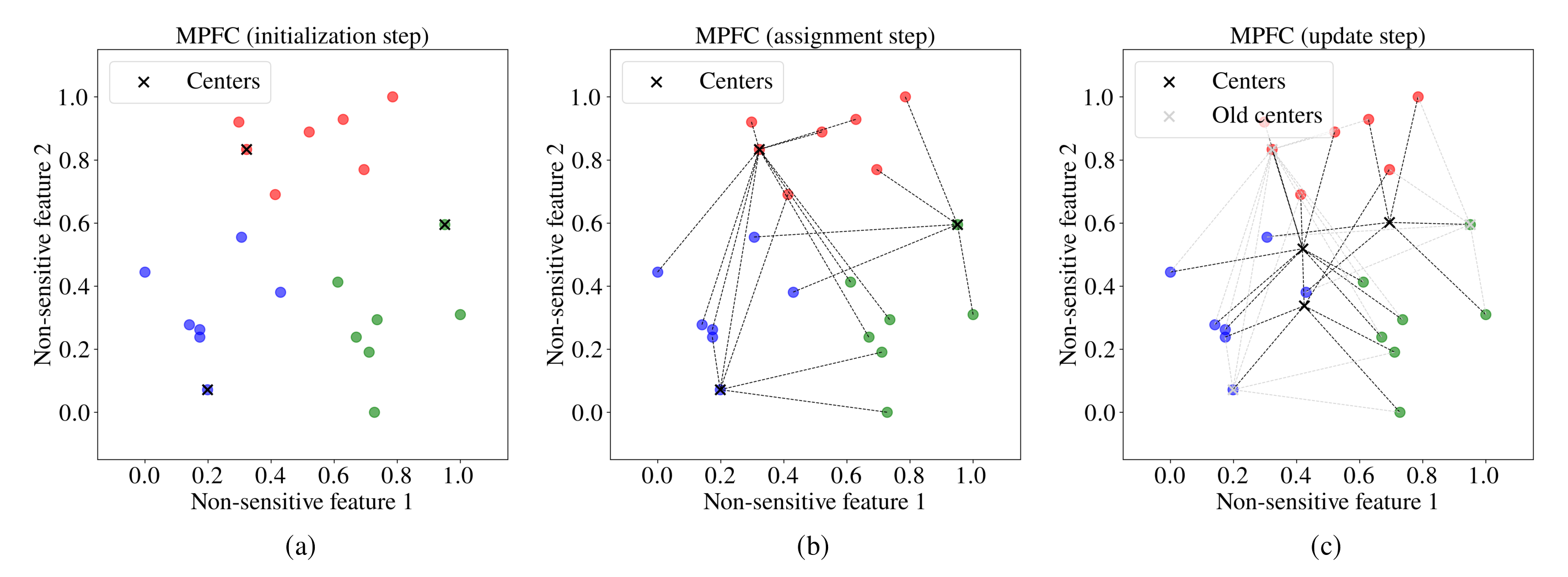}
    \caption{Applying MPFC to the illustrative example: (a) initialization of cluster centers; (b) assignment of objects; (c) update of cluster centers.}
    \label{fig:mpfc_example}
\end{figure}

Fig.~\ref{fig:mpfc_example} visualizes intermediate results from the first iteration of MPFC when applied to the illustrative example from Section~\ref{subsec:illustrative_example}. In Fig.~\ref{fig:mpfc_example} (a), three objects are selected as initial cluster centers using $k$-means++. These centers are then fixed during the assignment step in Fig.~\ref{fig:mpfc_example} (b), where the first clusters are obtained by assigning objects to centers by solving problem (BLP). Finally, Fig.~\ref{fig:mpfc_example} (c) shows the update of centers. In this small instance, the algorithm terminates after a few iterations and provides the optimal solution.

\subsection{Multi-stage minimum-cost flow-based fair clustering (MS-FlowFC)}\label{subsec:ms_flowfc}

The scalability of the MPFC algorithm is limited because the size of problem (BLP) grows rapidly with the number of objects $n$ and clusters $k$. As a result, solving it at each iteration becomes computationally expensive for large-scale instances. To address this limitation, we propose a more scalable heuristic referred to as multi-stage minimum-cost flow-based fair clustering (MS-FlowFC). Like the previous heuristic MPFC, MS-FlowFC also uses the decomposition scheme described in Section~\ref{subsec:decomposition_scheme}, but it differs in how it performs the assignment step. The key idea is to assign the objects to centers by solving multiple minimum-cost flow problems instead of solving one binary linear program. Minimum-cost flow problems are solvable in polynomial time, and there are highly optimized solution algorithms available (see \citealt{ortools}). Performing the assignment step by solving minimum-cost flow problems requires decomposing the assignment into multiple stages, where a specific protected group is assigned in a given stage. However, this decomposition is only possible when the problem instance has a single sensitive feature, as we describe later. Next, we provide a high-level description of the flow-based object assignment step, then describe the underlying flow network with the corresponding parameters, and finally illustrate the approach based on our illustrative example. 

As described above, the assignment step of MS-FlowFC is decomposed into multiple stages. In each stage, all objects of a specific protected group are assigned to the centers. The positions of the cluster centers remain fixed during all stages of the assignment step. In the first stage, the objects of the largest protected group are assigned to their nearest center. It is possible that some centers have no assignments. In this case, the positions of the centers with no assignments are re-initialized using $k$-means++ and the first stage is repeated until all centers have at least one object assigned. In the second stage, the objects from the second-largest protected group are assigned. Assigning every object from the second-largest protected group to its nearest center may violate the fairness constraints because of the assignments from the first stage. Therefore, cluster-specific lower and upper bounds need to be derived from the assignments of the first stage. The assignment problem can then be formulated as a minimum-cost flow problem, where the bounds are incorporated through appropriate node demands and arc capacities in the flow network. In the subsequent stages, the remaining protected groups are assigned similarly to the second stage in descending order of their size. When computing the lower and upper bounds on the number of objects that can be assigned to each cluster, the assignments of all preceding stages are considered. When all protected groups have been assigned, the MS-FlowFC heuristic continues with the update step of the decomposition scheme (see Fig.~\ref{fig:decomposition_scheme}). Note that MS-FlowFC assigns the protected groups in descending order of their size because no minimum-cost flow problem needs to be solved in the first stage, which offers a computational advantage in terms of running time. However, any sequence for the assignment of the protected groups would be possible. 

Next, we explain how the lower and upper bounds on the number of objects that are assigned to the centers in stage $l>1$ are derived. For simplicity, we use the index $l$ to denote both the stage and the protected group assigned in that stage, i.e., $g=l$. The lower bound on the number of objects from protected group $l$ that must be assigned to center $j$ depends on the number of objects of the protected groups that have been assigned in the preceding stages $l'=1, \ldots, l-1$. For each previously assigned protected group $l'$, a lower bound can be computed by multiplying the target balance $\overline{B}_s(X, \lambda)$ by the number of objects from this protected group assigned to center $j$, namely $|C_j \cap G_{l^{\prime}s}|$. The largest of these lower bounds is the one that must be imposed for protected group $l$ and center $j$ in stage $l$. We denote this lower bound by $\text{LB}_{jl}$, which is defined as follows:
\begin{equation}
    \label{eq:lower_bound}
    \text{LB}_{jl} = \max_{l^{\prime}=1,\dots,l-1}  \overline{B}_s(X, \lambda) \ |C_j \cap G_{l^{\prime}s}|, \quad \forall j.
\end{equation}

Similarly, the upper bound on the number of objects from protected group $l$ that can be assigned to center $j$ can be derived based on the least represented protected group in cluster $j$. We denote this upper bound by $\text{UB}_{jl}$, which is defined as follows:
\begin{equation}
    \label{eq:upper_bound}
    \text{UB}_{jl} = \min_{l^{\prime}=1,\dots,l-1} \frac{1}{\overline{B}_s(X, \lambda)} \ |C_j \cap G_{l^{\prime}s}|, \quad \forall j.
\end{equation}

\begin{figure}[t]
    \centering
    \includegraphics[width=0.95\textwidth]{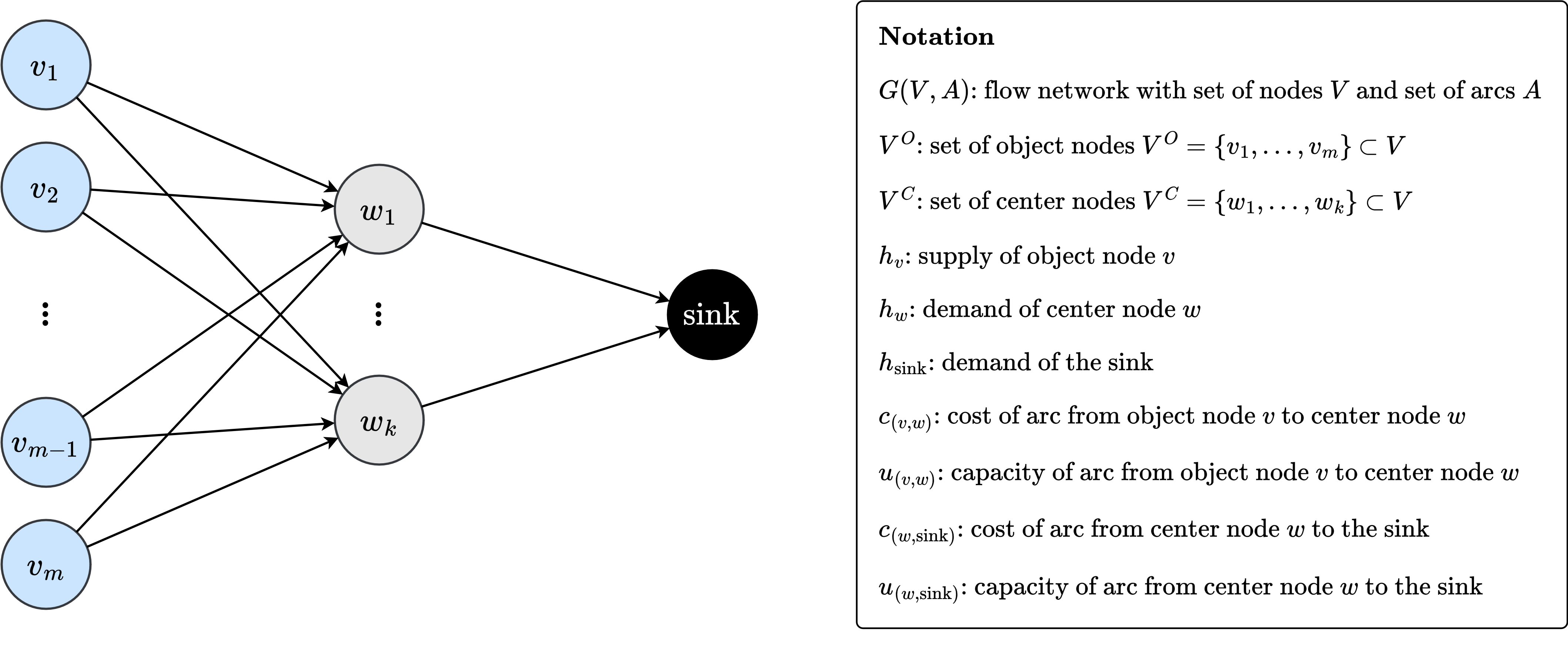}
    \caption{Flow network used in each stage $l>1$ of the assignment step of MS-FlowFC.}
    \label{fig:flow_network}
\end{figure}

Next, we describe the flow network that is constructed in stage $l > 1$ (see Fig.~\ref{fig:flow_network}). The flow network is modeled as a directed graph $G=(V, A)$, where $V$ is the set of nodes and $A$ is the set of arcs. Each node is associated with a demand or a supply and each arc is associated with a cost and a capacity. The set of nodes $V$ comprises the object nodes $V^O = \{v_1,\ldots,v_m\} \subset V$, the center nodes $V^C = \{w_1,\ldots,w_k\} \subset V$, and a sink node. The object nodes, colored blue in Fig.~\ref{fig:flow_network}, represent the $m$ objects of protected group $l$, and each object node $v \in V^O$ has a supply of $h_{v} = 1$. The center nodes, colored gray in Fig.~\ref{fig:flow_network}, represent the $k$ centers, and each center node $w \in V^C$ has a demand $h_{w}$, which we define below. The sink node, colored black in Fig.~\ref{fig:flow_network}, has a demand $h_{\text{sink}}$ set so that $\sum_{v \in V^O} h_v + \sum_{w \in V^C} h_w + h_{\textup{sink}} = 0$, ensuring that the flow network is balanced. Every object node $v \in V^O$ is connected to every center node $w \in V^C$ by an arc $(v,w) \in A$. Each of these arcs has a cost $c_{(v,w)}$ equal to the squared Euclidean distance between the object represented by $v$ and the center represented by $w$, and has capacity $u_{(v,w)} = 1$. Every center node $w \in V^C$ is connected to the sink by an arc $(w,\text{sink}) \in A$ that has cost $c_{(w,\text{sink})}=0$ and capacity $u_{(w,\text{sink})}$, which we describe below. 

By specifying the demand $h_w$ of a center node, we control the minimum number of objects that must be assigned to this center. By specifying the capacity $u_{(w,\text{sink})}$ of the arc that connects a center node to the sink, we control the maximum number of objects that can be assigned to this center. Together, the network parameters $h_w$ and $u_{(w,\text{sink})}$ form a range for the number of objects that can be assigned to each center node $w \in V^C$. To ensure the fairness requirements, this range is derived from the bounds $\text{LB}_{jl}$ in Eq.~\eqref{eq:lower_bound} and $\text{UB}_{jl}$ in Eq.~\eqref{eq:upper_bound}. Since all network parameters must be integer and the bounds may be fractional, we cannot directly use the bounds. To ensure that the fairness constraints are not violated, one could simply set the demands equal to the corresponding lower bounds rounded up and the capacities equal to the corresponding upper bounds rounded down minus the respective rounded demands. However, this simple rounding strategy may lead to infeasibility in two ways. First, setting the demands equal to the rounded-up lower bounds can lead to an unbalanced flow network, since the sum of the demands may no longer match the sum of the supplies. Second, setting the capacities according to the simple rounding strategy may prevent the demand of the sink from being satisfied, since the total capacity of the arcs from the center nodes to the sink can be insufficient to meet the demand of the sink. Conversely, using a rounding strategy that rounds down all lower bounds and rounds up all upper bounds is also not ideal, since it can potentially relax the fairness requirements for all centers and thereby lead to violations of the target balance. To prevent infeasibility while limiting the violations of the fairness requirements, we propose the following iterative parameter adjustment procedure. If the minimum-cost flow problem is infeasible, the difference between the total demand of the center nodes and the total supply is computed. If the total demand of the center nodes does not match the total supply, we apply a demand-adjustment procedure until the two are equal. In each iteration of this procedure, we estimate, for each center, the extent to which reducing its demand by one unit may violate the target balance. The demand of the center with the smallest potential violation is then reduced by one unit. Next, a capacity-adjustment procedure is applied until the total capacity of the arcs connecting the center nodes to the sink matches the demand of the sink. In each iteration of this procedure, we estimate, for each center, the extent to which increasing the capacity of the arc connecting it to the sink may violate the target balance. The capacity of the arc with the smallest potential violation is then increased by one unit. Using this adjustment procedure guarantees that an assignment can always be found, although the target balance may be violated. However, in instances where $n \gg k$, such violations are typically negligible, as shown in our computational analysis.

The adjustment of the demand of the center nodes is denoted by $\varepsilon_w \in \mathbb{Z}_{\geq0}$, while the adjustment of the capacity of the arcs from the center nodes to the sink is denoted by $\varepsilon_{(w,\text{sink})} \in \mathbb{Z}_{\geq0}$. The demands of the center nodes, adjusted for feasibility, are expressed as follows:
\begin{equation}
    \label{eq:demand}
    h_{w_j} = - \Bigl\lceil \text{LB}_{jl} \Bigr\rceil + \varepsilon_{w_j}, \quad \forall j.
\end{equation}

The capacities of the arcs from the center nodes to the sink, adjusted for feasibility, are defined as follows:

\begin{equation}
    \label{eq:capacity}
    u_{(w_j,\text{sink})} = \Bigl\lfloor \text{UB}_{jl} \Bigr\rfloor + h_{w_j} + \varepsilon_{(w_j,\text{sink})}, \quad \forall j.
\end{equation}

Note that $\varepsilon_{w_j}$ and $\varepsilon_{(w_j,\text{sink})}$ are initially set to zero and are increased only when required by the parameter adjustment procedure. 

\begin{figure}[t]
    \centering
    \includegraphics[width=1\textwidth]{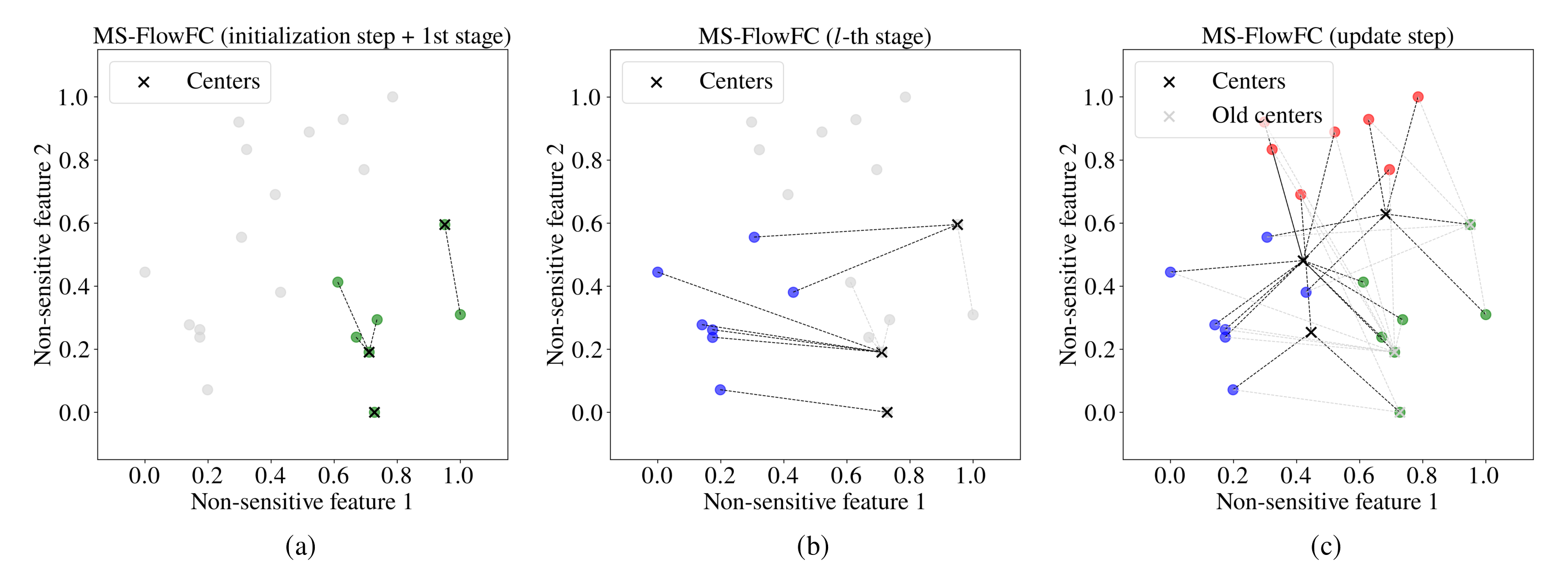}
    \caption{Applying MS-FlowFC to the illustrative example: (a) initialization of cluster centers and assignment of the first protected group; (b) assignment of the $l$-th protected group; (c) update of cluster centers.}
    \label{fig:msflowfc_example}
\end{figure}

Fig.~\ref{fig:msflowfc_example} visualizes intermediate results from the first iteration of MS-FlowFC when applied to the illustrative example from Section~\ref{subsec:illustrative_example}. In Fig.~\ref{fig:msflowfc_example} (a), three objects from the first protected group are selected as initial centers using $k$-means++, and all objects from the same protected group are assigned to their nearest center in the first stage. Fig.~\ref{fig:msflowfc_example} (b) shows the $l$-th stage, where objects from protected group $g=l=2$ are assigned to cluster centers by solving the corresponding minimum-cost flow problem. Finally, Fig.~\ref{fig:msflowfc_example} (c) illustrates the update of the centers obtained by aggregating the assignments from all previous stages.

\subsection{Scalable mathematical programming-based fair clustering (S-MPFC)}\label{subsec:s_mpfc}

The flow-based approach MS-FlowFC presented in the previous section is more scalable than MPFC but less general, as it is restricted to instances with a single sensitive feature. Moreover, for very large-scale datasets, solving the resulting minimum-cost flow problems can still be computationally demanding. Therefore, in this section we present a third heuristic that is as general as MPFC and even more scalable than MS-FlowFC. Like the previous approaches, it employs the decomposition scheme introduced in Section~\ref{subsec:decomposition_scheme}. We refer to this heuristic as the scalable mathematical programming-based fair clustering (S-MPFC) algorithm, as it can be viewed as a variant of MPFC. The idea of S-MPFC is to combine a modified version of MPFC with a preprocessing procedure that groups similar objects and represents each group by a representative. These representatives form a reduced dataset, which is then clustered using a modified version of MPFC. Finally, the labels of the representatives are mapped back to the original objects in the dataset. The groups of similar objects, referred to as \emph{batches}, are obtained by applying a vanilla clustering algorithm. Our approach is inspired by existing methods that employ preprocessing techniques, such as the approach of \cite{chierichetti2017fair}, which uses fairlet decomposition, and methods based on coreset construction (see \citealt{schmidt2019fair, huang2019coresets, bandyapadhyay2024coresets}). Our preprocessing technique offers advantages over fairlet decomposition and coreset construction. In contrast to fairlets, which must themselves satisfy fairness constraints, and coresets, which contain objects from only a single protected group, our batches impose no restrictions on their composition. Next, we provide a more detailed description of the preprocessing technique and then explain how the assignment and update steps of S-MPFC are implemented. 

The preprocessing technique starts by applying vanilla $k$-means clustering to the full dataset to identify $r$ clusters, where $k \leq r \leq n$. For clarity, we refer to these clusters as \emph{batches}, indexed by $b = 1, \dots, r$. Each batch $R_b$ is represented by its center, denoted by the vector $\vect{x}_b$, which is computed as the mean of the non-sensitive feature values of all objects in the batch. Furthermore, for each batch, we compute the weights $w_{bgs}$, defined as the number of objects in batch $R_b$ that belong to protected group $g$ of sensitive feature $s$. These weights are defined as follows:
\begin{equation}
    \label{eq:weights}
    w_{bgs} = |R_b \cap G_{gs}|, \quad (b=1,\ldots,r;\ g \in \mathcal{G}_{s};\ s \in S).
\end{equation}

In the assignment step, the binary linear programming formulation (BLP$'$) is solved to assign the representatives to cluster centers. The formulation (BLP$'$) differs from the formulation (BLP) used in MPFC primarily in that it incorporates the weights of the representatives into the fairness constraints. In formulation (BLP$'$), we use parameter $c_{bj}$ to denote the squared Euclidean distance between representative $b$ and center $j$. The decision variable $x_{bj}$ is equal to one if representative $b$ is assigned to center $j$, and zero otherwise. Formulation (BLP$'$) is given as follows:
\begin{empheq}[left=\text{(BLP$'$)}\empheqlbrace]{align}
  \min \quad & \sum_{b=1}^{r} \sum_{j=1}^k c_{bj} x_{bj} \label{s_mpfc_a}\\
  \text{s.t.} \quad & \sum_{j=1}^k x_{bj} = 1 && (b=1,\ldots,r) \label{s_mpfc_b}\\
  & \sum_{b=1}^{r} x_{bj} \geq 1 && (j=1,\ldots,k) \label{s_mpfc_c}\\
  & \sum_{b=1}^{r} w_{bgs} x_{bj} \geq \overline{B}_s(X, \lambda) \sum_{b=1}^{r} w_{bg^{\prime}s} x_{bj} && (g,g' \in \mathcal{G}_s: g \ne g';\ j=1,\ldots,k;\ s \in S) \label{s_mpfc_d}\\
  & x_{bj} \in \{0, 1\} && (b=1,\ldots,r;\ j=1,\ldots,k) \label{s_mpfc_e}
\end{empheq}

The objective function (\ref{s_mpfc_a}) minimizes the total squared Euclidean distance between representatives and cluster centers. Constraints (\ref{s_mpfc_b}) and (\ref{s_mpfc_c}) ensure that each representative is assigned to exactly one cluster and that no cluster remains empty, respectively. Constraints (\ref{s_mpfc_d}) enforce fairness by taking into account the weights of the representatives. By considering the weights of the representatives, it is guaranteed that the target balance is also achieved after mapping the labels from the representatives back to the original objects in the dataset. Constraints (\ref{s_mpfc_e}) define the domain of the decision variables.

In the update step, the positions of the cluster centers $\vect{\mu}_j$ are recomputed as the weighted mean of the representatives assigned to each cluster. Specifically, each representative $\vect{x}_b$ is weighted by $|R_b|$, the number of objects in its batch, so that representatives corresponding to larger batches have a greater influence on the computation of the updated cluster center. The updated cluster centers are computed as follows:
\begin{equation}
    \label{eq:weighted_update}
    \vect{\mu}_j = \frac{\displaystyle \sum_{b=1}^{r} |R_b| \, x_{bj} \, \vect{x}_b}{\displaystyle \sum_{b=1}^{r} |R_b| \, x_{bj}}, \quad (j=1,\dots,k).
\end{equation}

\begin{figure}[t]
    \centering
    \includegraphics[width=1\textwidth]{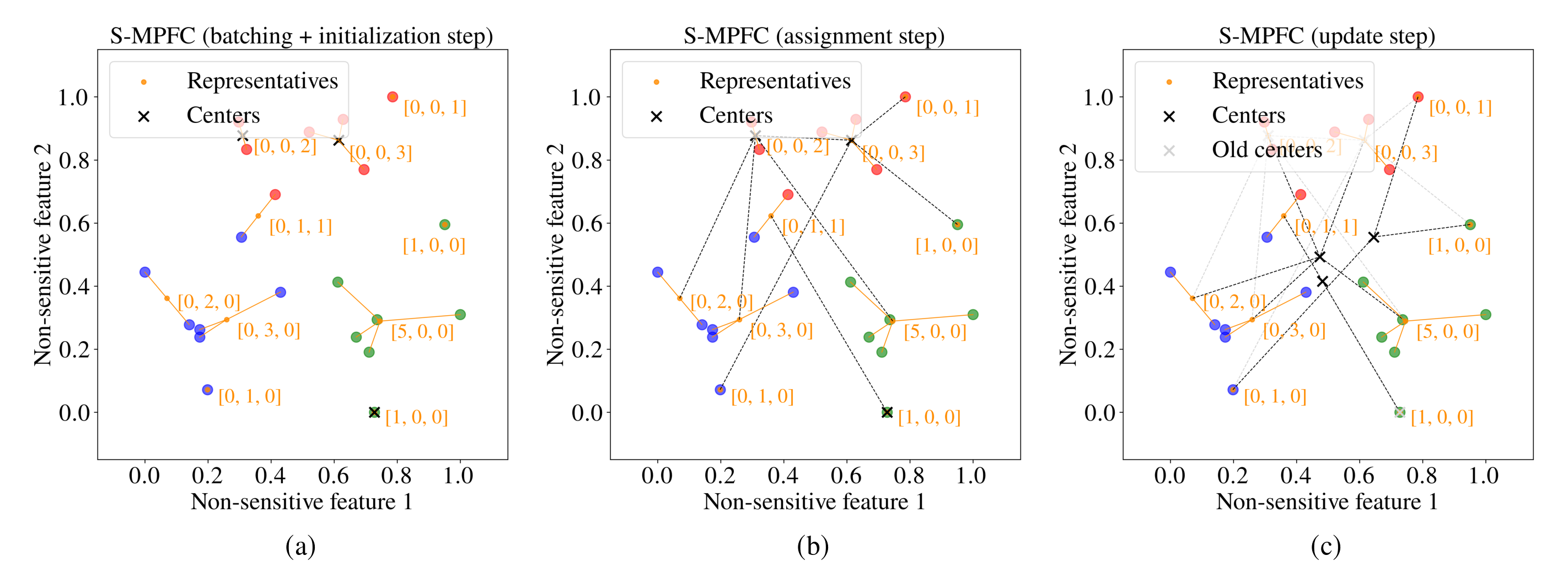}
    \caption{Applying S-MPFC to the illustrative example: (a) generation of $r=10$ batches and initialization of cluster centers; (b) assignment of representatives; (c) update of cluster centers.}
    \label{fig:smpfc_example}
\end{figure}

Fig.~\ref{fig:smpfc_example} visualizes intermediate results from the first iteration of S-MPFC when applied to the illustrative example from Section~\ref{subsec:illustrative_example}. Fig.~\ref{fig:smpfc_example} (a) shows the preprocessing step, where the dataset is partitioned into $r=10$ batches. Each batch has a representative, indicated by an orange dot and computed as the mean of the feature values of its objects. The orange lines depict the assignment of objects to their respective representatives. Each representative has three weights (reported in brackets), indicating the number of green, blue, and red objects included in the corresponding batch. Additionally, in Fig.~\ref{fig:smpfc_example} (a), three representatives are selected as initial centers using $k$-means++. Following the decomposition scheme described earlier, Fig.~\ref{fig:smpfc_example} (b) and Fig.~\ref{fig:smpfc_example} (c) show the first assignment of representatives and the subsequent update of centers, respectively. Upon termination of the algorithm, the assignment of representatives to centers is used to derive the assignment of the original objects to the same centers.

\section{Numerical experiments}\label{sec:numerical_experiments}

In this section, we present the numerical experiments conducted to evaluate the performance of the proposed heuristics and compare it to the performance of benchmark methods. Section~\ref{subsec:benchmark_methods} presents the benchmark methods, Section~\ref{subsec:data} describes the datasets used, and Section~\ref{subsec:experimental_design} outlines the experimental design. Sections~\ref{subsec:small_results}, \ref{subsec:medium_results}, and \ref{subsec:large_results} report and discuss the numerical results obtained for small-, medium-, and large-scale datasets, respectively. Finally, Section~\ref{subsec:tradeoff_results} analyzes how effectively the proposed methods and a benchmark method control the trade-off between clustering cost and fairness.

Our heuristics are implemented in Python 3.12. Gurobi Optimizer 12 is used to solve the binary linear programming problem arising in the assignment step of MPFC and S-MPFC. The Faiss library \citep{11202651} is employed in the first stage of the assignment step of MS-FlowFC and in the preprocessing step of S-MPFC. Google OR-Tools is used to solve the minimum-cost flow problems in the assignment step of MS-FlowFC. The source code of our heuristics is publicly available in a GitHub$^{\ref{github:code}}$ repository. The repository includes instructions for reproducing the reported results and installing the software as a Python package. 

\subsection{Benchmark methods}\label{subsec:benchmark_methods}

In our experiments, we evaluate the proposed heuristics against five benchmark methods: two methods from the literature, two exact methods introduced in Appendix \ref{app:exact}, and a vanilla clustering algorithm. The two benchmark methods from the literature are the scalable fair clustering algorithm proposed by \cite{backurs2019scalable}, denoted here by SFC, and the order-and-cut algorithm proposed by \cite{li2024optimization}, denoted here by O\&C. The two exact methods are a mixed-integer quadratically constrained programming formulation solved with Gurobi Optimizer 12, denoted here by MIQCP, and a set-variable-based formulation solved with Hexaly 14, denoted here by SetVars. Exact methods provide certified optimal solutions that are effective reference points for assessing heuristics, as highlighted by \citet{piccialli2022sos}. The vanilla clustering algorithm is Lloyd’s algorithm introduced in \cite{lloyd1982least}, denoted here by Lloyd. Lloyd’s algorithm does not consider fairness and therefore serves as a baseline for evaluating the impact of fairness constraints on clustering cost. Next, we provide a more detailed description of the two benchmark methods from the literature.

The SFC algorithm proposed by \cite{backurs2019scalable} is widely used in the fair clustering literature as a benchmark method, particularly for large-scale fair clustering applications, such as the cybersecurity application described in \cite{sahoo2022evaluation}. The SFC algorithm consists of two stages. In the first stage, the dataset is partitioned into groups of objects that satisfy the fairness constraints. These groups, referred to as fairlets, are generated using the $(p,q)$-fairlet decomposition. The $(p,q)$-fairlet decomposition takes as input two integers $p$ and $q$ with $1\leq p\leq q$ and ensures that each resulting fairlet has a balance of at least $\frac{p}{q}$. In the second stage, a vanilla clustering algorithm is applied to cluster the fairlets. 

Since the $(p, q)$-fairlet decomposition only applies to instances with a single sensitive feature and two protected groups, we used the SFC algorithm as a benchmark only for such instances. The SFC algorithm also has two additional limitations. The first additional limitation is that the clustering cost within fairlets increases as $p$ and $q$ grow. We also observed empirically in our experiments that the running time of the $(p,q)$-fairlet decomposition increases with increasing values of $p$ and $q$. The second additional limitation is that for some values of $p$ and $q$, the $(p, q)$-fairlet decomposition may generate a small number of fairlets, resulting in poor-quality solutions or, if fewer fairlets are generated than there are clusters, infeasibility. While we cannot address the second limitation, we address the first limitation as follows. We approximate the target balance $\overline{B}_s(X, \lambda)$ using values of $p$ and $q$ less than or equal to 1{,}000, following the procedure described in Appendix \ref{app:algorithm}. In cases where it is not possible to obtain a ratio $\frac{p}{q}$ with $1 \leq p \leq q \leq 1000$ that exactly matches the target balance, the procedure described in Appendix \ref{app:algorithm} returns values for $p$ and $q$ that yield a slightly smaller ratio. This procedure gives SFC a slight advantage, since a lower target balance can lead to reduced clustering cost. However, in our experiments, this effect was negligible. SFC was originally proposed for the $k$-median objective. In our study, we adapt its second stage by using the $k$-means objective instead. Because the fairlet decomposition in SFC relies on a randomly initialized seed, different solutions can be obtained by changing this seed. In our experiments, we therefore ran SFC multiple times with different random seeds and reported the best solution found as well as the total running time over all runs. The SFC algorithm is implemented in Python~3.6 and MATLAB, and its source code is publicly available. 

The O\&C algorithm proposed by \cite{li2024optimization} is a recent approach to explicitly optimize the trade-off between clustering cost and fairness. The O\&C algorithm consists of an ordering step, which puts all objects into a sequence, and a cutting step, which partitions the sequence into clusters. The goal of the ordering step is to produce a sequence of objects that, when partitioned in the subsequent cutting step, results in clusters with both low clustering cost and the desired level of fairness. More specifically, in the ordering step, three orderings of the objects are constructed. In the first ordering, similar objects are placed close to each other. This ordering is obtained by combining vanilla clustering with dimensionality reduction. Since cutting such an ordering will result in clusters with low clustering cost, we refer to it as the cost-optimized ordering. In the second ordering, objects are placed such that blocks of consecutive objects are fair with respect to the protected groups. This ordering is constructed by adjusting the cost-optimized ordering. Since cutting the second ordering will result in fair clusters, we refer to it as the fairness-optimized ordering. The third ordering is a balanced ordering between the cost-optimized and the fairness-optimized orderings. This balanced ordering is obtained by using a logistic transition function that interpolates between the cost-optimized and the fairness-optimized ordering. A parameter $\alpha \geq 0$ of the logistic transition function influences the fairness level of the balanced ordering. Higher values of $\alpha$ place more emphasis on the fairness-optimized ordering and thus tend to increase fairness in the balanced ordering. In the cutting step, optimal positions to cut the balanced ordering are identified by solving a shortest path problem on an auxiliary graph. The O\&C algorithm is implemented in MATLAB, and its source code is publicly available. 

In contrast to the SFC algorithm, the O\&C algorithm can be applied to instances where a single sensitive feature has more than two protected groups. This allows us to use it as a benchmark approach for instances where the SFC~algorithm is not applicable. However, like the SFC algorithm, the O\&C algorithm is not applicable to instances with multiple sensitive features. For this reason, we did not consider such instances in our experiments. The O\&C algorithm has two additional limitations that affect our experiments. First, its limited scalability restricts comparisons to small instances with up to 5{,}000 objects. Second, the parameter $\alpha$ provides less precise control over the trade-off between cost and fairness than the parameter $\lambda$ used in the proposed heuristics. While $\lambda$ directly determines the target balance, there is no direct relationship between $\alpha$ and the target balance. To ensure a fair comparison between the O\&C algorithm and our heuristics, we first applied the O\&C algorithm using a fixed value of $\alpha$ and then used the resulting clustering balance as the target balance for our heuristics.

\subsection{Data}\label{subsec:data}
We used two small-scale synthetic datasets (\syntha\ and \synthb) introduced in this paper, five medium- to large-scale datasets from the UCI Machine Learning Repository (see \citealt{ucimlrepo}) that are commonly used in the fair clustering literature (\credit, \bank, \adult, \diabetes, and \census; see \citealt{chhabra2021overview}), and a large-scale dataset from the Centre for Research in Cyber Security at the Singapore University of Technology and Design (\swat; see \citealt{itrust}). For each dataset, Table~\ref{tab:datasets} summarizes the dataset size, the number of non-sensitive features, the sensitive feature, and the corresponding protected groups. Next, we briefly describe each dataset.

In the datasets \syntha\ and \synthb, the objects represent points in a two-dimensional space. The two datasets differ in how the points are generated. In \syntha, all points are sampled from a single Gaussian distribution. In \synthb, a separate Gaussian distribution is used for each protected group. In the \credit\ dataset, the objects represent credit card holders of a Taiwanese bank \citep{default_of_credit_card_clients_350}. In the \bank\ dataset, the objects are records from phone calls conducted during a marketing campaign of a Portuguese retail bank \citep{bank_marketing_222}. In the \adult\ dataset, the objects are individuals sampled from the 1994 United States census \citep{adult_2}. In the \diabetes\ dataset, the objects are hospitalized diabetic patients from 130 hospitals in the United States \citep{diabetes_34}. In the \census\ dataset, the objects are individuals sampled from the 1990 United States census \citep{us_census_data_(1990)_116}. Finally, in the Secure Water Treatment (\swat) dataset, the objects are measurements from sensors of a water treatment plant operating under both normal conditions and during cyber attacks \citep{goh2016swat, itrust}. \citet{sahoo2022evaluation} and \citet{handa2022evaluating} used this dataset to evaluate the scalable fair clustering method of \cite{backurs2019scalable} for cyber-physical system security.

\begin{table}[t]
    \caption{Datasets used in the numerical experiments.}
    \label{tab:datasets}
    \footnotesize
    \centering
    \setlength{\tabcolsep}{15.5pt}
    \renewcommand{\arraystretch}{1.2}
    \begin{tabular}{l r r l l}
        \toprule
        Dataset & $n$ & $d$ & Sensitive feature & Protected groups (size) \\ 
        \midrule
        \syntha & 21 & 2 & \texttt{color} & Green (7), Blue (7), Red (7) \\
        \synthb & 21 & 2 & \texttt{color} & Green (7), Blue (7), Red (7) \\
        \bankfive$^{*}$ & 5,000 & 7 & \texttt{marital} & \makecell[t]{Married (3{,}010), Single (1{,}415), Divorced (575)} \\
        \credit & 30,000 & 22 & \texttt{X2} & Male (11{,}888), Female (18{,}112) \\
        \bankforty$^{*}$ & 40,004 & 7 & \texttt{marital} & Married (27{,}214), Single (12{,}790) \\
        \adult & 48,842 & 6 & \texttt{sex} & Male (32{,}650), Female (16{,}192) \\
        \diabetes & 101,763 & 11 & \texttt{gender} & Male (47{,}055), Female (54{,}708) \\
        \swat & 944,882 & 51 & \texttt{Normal/Attack} & Normal (890{,}298), Attack (54{,}584) \\
        \census & 2,458,285 & 25 & \texttt{iSex} & Male (1{,}266{,}684), Female (1{,}191{,}601) \\
        \midrule
        \multicolumn{5}{l}{$^{*}$ derived from the \bank\ dataset} \\
        \bottomrule
    \end{tabular}
\end{table}

We preprocessed the downloaded datasets as follows. First, we separated the sensitive feature from the remaining features, since the clustering cost is computed only on the non-sensitive features. We then applied min-max scaling to normalize each non-sensitive feature to the range [0, 1]. For each dataset, we selected a sensitive feature that has been used in prior fair clustering studies. For \credit, we used \texttt{X2} as the sensitive feature, which represents the gender of a client. For \bank, we used \texttt{marital} as the sensitive feature, which represents the marital status of a client. For \adult, we used \texttt{sex} as the sensitive feature, which represents the gender of a person. For \diabetes, we used \texttt{gender} as the sensitive feature, which represents the gender of a patient. For \swat, we used \texttt{Normal/Attack} as the sensitive feature, which represents whether the system is under attack or not. For \census, we used \texttt{iSex} as the sensitive feature, which represents the gender of a person. As done by \cite{backurs2019scalable}, we kept only the first 25 non-sensitive features for \census. The protected groups of these sensitive features are listed in Table~\ref{tab:datasets}.

The \bank\ dataset contains $n=45{,}211$ objects, and its sensitive feature \texttt{marital} comprises three protected groups (Married, Single, and Divorced). Since the benchmark approach SFC of \cite{backurs2019scalable} can only handle two protected groups, we constructed the \bankforty\ dataset by removing the $5{,}207$ objects of \bank\ belonging to the smallest protected group (Divorced), resulting in a dataset with only two protected groups. Since the benchmark approach O\&C of \cite{li2024optimization} cannot be applied to the original dataset within a reasonable running time, we constructed the \bankfive\ dataset by applying stratified sampling to \bank, resulting in a subsampled version with $5{,}000$ objects in which the distribution of the three protected groups is approximately preserved. 

\subsection{Experimental design}\label{subsec:experimental_design}

We designed four numerical experiments to compare the proposed heuristics (MPFC, S-MPFC, and MS-FlowFC) with the benchmark methods (Lloyd, MIQCP, SetVars, SFC, O\&C) across different datasets (small-, medium-, and large-scale), numbers of clusters, and fairness levels. The following performance metrics are used in the comparison: clustering cost, clustering balance, running time, and, for the exact methods, the MIP Gap. In all four experiments, we repeatedly applied the approaches that rely on random initialization, namely MPFC, MS-FlowFC, S-MPFC, SFC, and Lloyd, with different random seeds until either a time limit was reached or 100 runs had been completed. We then reported the best solution found and the total running time across all runs performed. Note that for S-MPFC, we used the same representatives for each run, i.e., we applied the preprocessing technique only once. For the proposed heuristics, we set the minimum improvement to $\Delta=0.1\%$ and the maximum number of iterations per run to $\texttt{MaxIter}=100$. However, $\texttt{MaxIter}$ was never reached in our experiments. All computational experiments were conducted on an HP workstation equipped with an Intel(R) Xeon(R) Silver 4114 CPU (2.20 GHz, 10 cores) and 128 GB of RAM.

In the first experiment, we investigate whether the proposed heuristics MPFC and MS-FlowFC are able to construct optimal solutions. To obtain certified optimal solutions, we used the two exact approaches MIQCP and SetVars. Since obtaining certified optimal solutions within reasonable running time is feasible only for small-scale datasets, we restricted this experiment to the datasets \syntha\ and \synthb. In this experiment, we also applied Lloyd's algorithm to assess the increase in clustering cost when incorporating fairness. For each dataset, we generated four problem instances by varying the number of clusters $k \in \{2, 3, 4, 5\}$. Since, in both datasets, all protected groups are of equal size and the parameter $k$ is smaller than the size of each protected group, the dataset balance $B_s(X)$ can be achieved for all values of $k$. We set the tolerance parameter $\lambda=0.00$ and imposed a time limit of $\texttt{MaxTime}=3{,}600$ seconds for each instance.

In the second experiment, we compare the performance of the proposed heuristics MPFC and MS-FlowFC with the benchmark method SFC on the well-known datasets \credit, \bankforty, \adult, and \diabetes\ from the literature. The benchmark method O\&C was not included in this experiment because it was unable to produce solutions within acceptable running times for these datasets. In addition to SFC, we also applied the exact approaches MIQCP and SetVars, and the vanilla clustering approach Lloyd to obtain reference solutions. For each dataset, we generated twelve problem instances by varying the number of clusters $k \in \{2, 5, 10, 15\}$ and the values of the tolerance parameter $\lambda \in \{0.01, 0.10, 0.30\}$. As mentioned in Section~\ref{subsec:problem_description}, it is possible that the dataset balance $B_s(X)$ cannot be achieved for some values of $k$. For example, consider the illustrative example from Section~\ref{subsec:illustrative_example}, where each of the three protected groups contains seven objects. In this case, the dataset balance is equal to one. For specific values of the parameter $k$, such as $k=8$, no partition exists that achieves the dataset balance. Therefore, in our experiments, we replaced the dataset balance $B_s(X)$ with a value $B_s(X, k)$ computed with Eq.~\eqref{eq:maximum_balance}, which takes into account the sizes of the protected groups and the parameter $k$.
\begin{equation}
    \label{eq:maximum_balance}
    B_s(X, k)=\frac{\displaystyle\Bigg\lfloor \frac{\min_g (|G_{gs}|)}{k} \Bigg\rfloor}{\displaystyle\Bigg\lceil \frac{\max_g (|G_{gs}|)}{k} \Bigg\rceil},
\end{equation}

The expressions $\min_g (|G_{gs}|)$ and $\max_g (|G_{gs}|)$ in Eq.~\eqref{eq:maximum_balance} denote the sizes of the smallest and largest protected groups, respectively. As in the first experiment, we also imposed a time limit of $\texttt{MaxTime} = 3{,}600$ seconds for each instance in the second experiment. 

In the third experiment, we compare the scalability of the proposed heuristics MS-FlowFC and S-MPFC with the scalability of the benchmark method SFC on the large-scale datasets \swat\ and \census. The control parameter $r$ of S-MPFC, which determines the size of the reduced dataset, was set to a small value ($r=100$) to ensure short running times. In cases where solution quality is more important than running time, we recommend increasing parameter $r$. As in the previous experiments, we also applied the vanilla clustering algorithm Lloyd to obtain reference solutions. We did not include the exact approaches and the O\&C approach, as they do not scale to datasets with close to or more than one million objects. Unlike in the second experiment, where the target balance was computed from $B_s(X,k)$ and the tolerance parameter $\lambda$, here we specify the target balance directly. We chose target balance values such that the benchmark approach SFC could represent them as “simple” ratios $\frac{p}{q}$ with small values of $p$ and $q$. As mentioned in Section~\ref{subsec:benchmark_methods}, we found that large values of $p$ and $q$ can significantly increase the running time of SFC. For the \swat\ dataset, we generated twelve problem instances by varying the number of clusters $k \in \{2, 5, 10, 15\}$ and the values of the target balance $\textup{Target} \in \{0.06, 0.04, 0.02\}$. For the \census\ dataset, we generated twelve problem instances by varying the number of clusters $k \in \{2, 5, 10, 15\}$ and the values of the target balance $\textup{Target} \in \{0.90, 0.75, 0.50\}$. As in the previous experiments, we also imposed a time limit of $\texttt{MaxTime}=3{,}600$ seconds for each instance in the third experiment. 

In the fourth experiment, we compare the performance of our heuristics MPFC and MS-FlowFC with the benchmark approach O\&C for different desired levels of fairness. In this experiment, we used the \bankfive\ dataset, following the experimental design of \cite{li2024optimization}, which also used a downsampled version of the original \bank\ dataset. As mentioned in Section~\ref{subsec:benchmark_methods}, our heuristics and the benchmark approach O\&C use different parameters to control the fairness level. Our heuristics employ parameter $\lambda$, while approach O\&C uses parameter $\alpha$. Since there is no direct correspondence between the two parameters, we first applied O\&C with a given value of $\alpha$, computed the clustering balance of the resulting solution, and then used this clustering balance as the target balance for our heuristics. We generated sixteen instances by varying the number of clusters $k \in \{2, 5, 10, 15\}$ and the values of parameter $\alpha \in \{0.0, 0.5, 1.5, 2.0\}$. Higher values of $\alpha$ tend to lead to higher levels of fairness. As in \citet{li2024optimization}, we used values of $\alpha \in [0, 2]$.  

In addition to comparing the performance of the benchmark approach O\&C and our heuristics MPFC and MS-FlowFC for different levels of fairness, we used the \bankfive\ dataset to evaluate how effectively the trade-off between clustering cost and fairness can be controlled by our heuristics compared to the benchmark approach. To this end, we generated eleven instances for each approach by varying the fairness level from low to high using the respective parameter (either $\lambda$ for MPFC and MS-FlowFC or $\alpha$ for O\&C). In the fourth experiment, we imposed an extended time limit of $\texttt{MaxTime} = 10{,}800$ seconds, as the benchmark approach O\&C could not find a feasible solution within one hour for the given instances.

\subsection{Results for small-scale instances}\label{subsec:small_results}

The results for the small-scale instances of datasets \syntha\ and \synthb\ are reported in Table~\ref{tab:results_synthetic_a} and Table~\ref{tab:results_synthetic_b}, respectively. Both tables contain one row for each instance. The first three columns state the number of clusters ($k$), the tolerance parameter ($\lambda$), and the target balance (Target) of each instance. Columns four to eight report the clustering costs for each method. Columns nine to thirteen report the total running time (in seconds) for each method. The clustering balance is not reported, as it always matches the target balance, except for solutions obtained with Lloyd, which does not account for fairness. As noted in Section~\ref{subsec:experimental_design}, the methods Lloyd, MPFC, and MS-FlowFC were each run 100 times with different random seeds. The reported clustering cost is the best value obtained across runs, and the reported running time is the total time over all 100 runs. For the two exact methods (MIQCP and SetVars), we list the relative optimality gap in parentheses after the clustering costs. We can draw the following main conclusions from Table~\ref{tab:results_synthetic_a} and Table~\ref{tab:results_synthetic_b}: 

\begin{itemize}[noitemsep, topsep=0pt]
    \item For six out of eight instances, a certified optimal solution is found. 
    \item For every instance where optimality is certified, the proposed heuristic MPFC finds an optimal solution. For the remaining two instances, it finds the best solution obtained with any fair clustering method. 
    \item MS-FlowFC (abbreviated as “Flow” in the table) achieves the best solution in five out of eight instances, three of which are certified optimal solutions.     
    \item MPFC and MS-FlowFC always complete 100 runs within a few seconds. 
\end{itemize}

Other interesting insights can be drawn from the tables. Although all instances used in this experiment comprise only 21 objects, the exact approaches are not always able to prove optimality within one hour. The running time and the relative optimality gap tend to increase with increasing values of the number of clusters ($k$). This highlights the need for fast and effective heuristics for fair clustering. Among the exact approaches, MIQCP is usually faster in proving optimality, while SetVars typically finds high-quality solutions earlier during the search. Comparing the clustering costs of the fair solutions with those of the vanilla clustering solutions reveals that the difference is dataset-dependent. For instances derived from \syntha, where the points (objects) of different protected groups are sampled from the same Gaussian distribution, the clustering costs differ only slightly. In contrast, for instances derived from \synthb, where the points (objects) of different protected groups are sampled from different Gaussian distributions, the difference in clustering cost is considerably larger. 

\subsection{Results for medium-scale instances}\label{subsec:medium_results}

The results for the medium-scale instances are consistent across the datasets \credit, \bankforty, \adult, and \diabetes. We therefore present here only the results for the instances derived from the \bankforty\ dataset. The results for the remaining datasets are available on GitHub\footnote{\label{github:results}\url{https://github.com/claudio-mantuano/fair-clustering-at-scale/tree/main/paper_results}}. 

\begin{table}[t]
    \caption{Results for the \syntha\ dataset.}
    \label{tab:results_synthetic_a}
    \footnotesize
    \setlength{\tabcolsep}{6pt}
    \begin{tabular}{rrrrrrrrrrrrr}
        \toprule
         & & & \multicolumn{5}{l}{Cost} & \multicolumn{5}{l}{CPU [s]} \\
        \cmidrule(lr){4-8}\cmidrule(lr){9-13}
        $k$ & $\lambda$ & Target 
        & $\text{Lloyd}^{*}$ & MIQCP (Gap) & SetVars (Gap) & MPFC & Flow 
        & $\text{Lloyd}^{*}$ & MIQCP & SetVars & MPFC & Flow \\
        \midrule
        2 & 0.00 & 1.00 & 1.756 & \textbf{1.948} (0.00) & \textbf{1.948} (0.29) & \textbf{1.948} & \underline{1.995} & 2.21 & 0.33 & limit & 2.60 & 1.00\\
        3 & 0.00 & 1.00 & 1.157 & \textbf{1.173} (0.00) & \textbf{1.173} (0.00) & \textbf{1.173} & \textbf{1.173} & 1.70 & 2.80 & 2266.35 & 3.65 & 1.08\\
        4 & 0.00 & 1.00 & 0.753 & \textbf{0.874} (0.00) & \textbf{0.874} (0.76) & \textbf{0.874} & \underline{1.010} & 1.69 & 30.76 & limit & 3.67 & 0.98\\
        5 & 0.00 & 1.00 & 0.509 & \textbf{0.721} (0.00) & \textbf{0.721} (0.72) & \textbf{0.721} & \underline{0.809} & 1.70 & 467.93 & limit & 4.37 & 0.97\\
        \midrule
        \multicolumn{13}{l}{$(^{*})$ vanilla clustering; \ (limit) time limit of 3{,}600 seconds reached; \ (\textbf{bold}) best fair result; \ (\underline{underline}) 2nd best fair result}\\
        \bottomrule
    \end{tabular}
\end{table}

\begin{table}[t]
    \caption{Results for the \synthb\ dataset.}
    \label{tab:results_synthetic_b}
    \footnotesize
    \setlength{\tabcolsep}{6pt}
    \begin{tabular}{rrrrrrrrrrrrr}
        \toprule
         & & & \multicolumn{5}{l}{Cost} & \multicolumn{5}{l}{CPU [s]} \\
        \cmidrule(lr){4-8}\cmidrule(lr){9-13}
        $k$ & $\lambda$ & Target 
        & $\text{Lloyd}^{*}$ & MIQCP (Gap) & SetVars (Gap) & MPFC & Flow 
        & $\text{Lloyd}^{*}$ & MIQCP & SetVars & MPFC & Flow \\
        \midrule
        2 & 0.00 & 1.00 & 1.905 & \textbf{3.013} (0.00) & \textbf{3.013} (0.00) & \textbf{3.013} & \textbf{3.013} & 2.16 & 1.31 & 523.07 & 4.33 & 0.74\\
        3 & 0.00 & 1.00 & 0.868 & \textbf{2.859} (0.00) & \textbf{2.859} (0.72) & \textbf{2.859} & \textbf{2.859} & 1.71 & 490.73 & limit & 6.26 & 0.89\\
        4 & 0.00 & 1.00 & 0.652 & \textbf{2.783} (0.51) & \textbf{2.783} (0.94) & \textbf{2.783} & \textbf{2.783} & 1.71 & limit & limit & 9.68 & 1.17\\
        5 & 0.00 & 1.00 & 0.468 & \textbf{2.723} (0.74) & \textbf{2.723} (1.00) & \textbf{2.723} & \textbf{2.723} & 1.69 & limit & limit & 10.87 & 1.27\\
        \midrule
        \multicolumn{13}{l}{$(^{*})$ vanilla clustering; \ (limit) time limit of 3{,}600 seconds reached; \ (\textbf{bold}) best fair result; \ (\underline{underline}) 2nd best fair result}\\
        \bottomrule
    \end{tabular}
\end{table}

\begin{table}[t]
    \caption{Clustering quality results for the \bankforty\ dataset.}
    \label{tab:bank_cost_results}
    \footnotesize
    \setlength{\tabcolsep}{5.1pt}
    \begin{tabular}{lrrrrrrcrrrrcrrr}
        \toprule
        & & & & & \multicolumn{2}{l}{Lloyd} & \multicolumn{5}{l}{Cost} & \multicolumn{4}{l}{$\text{Gap}_{\text{Cost}}^{\text{SFC}}$ [\%]} \\
        \cmidrule(lr){6-7}\cmidrule(lr){8-12}\cmidrule(lr){13-16}
        $k$ & $\lambda$ & Target & $p$ & $q$ & Cost & Bal. & MIQCP & SetVars & MPFC & Flow & SFC & MIQCP & SetVars & MPFC & Flow\\
        \midrule
        2 & 0.01 & 0.47 & 67 & 144 & 2,396 & 0.46 & - & \underline{2,399} & \textbf{2,397} & \textbf{2,397} & 2,426 & - & -1.12 & -1.18 & -1.18\\
        2 & 0.10 & 0.42 & 335 & 792 &  &  & - & \textbf{2,396} & \textbf{2,396} & \textbf{2,396} & \underline{2,536} & - & -5.50 & -5.50 & -5.50\\
        2 & 0.30 & 0.33 & 126 & 383 &  &  & - & \textbf{2,396} & \textbf{2,396} & \textbf{2,396} & \underline{2,410} & - & -0.56 & -0.56 & -0.56\\
        5 & 0.01 & 0.47 & 375 & 806 & 1,326 & 0.10 & - & 1,515 & \textbf{1,389} & \underline{1,402} & 1,831 & - & -17.26 & -24.15 & -23.43\\
        5 & 0.10 & 0.42 & 291 & 688 &  &  & - & 1,400 & \textbf{1,380} & \underline{1,390} & 1,718 & - & -18.47 & -19.67 & -19.09\\
        5 & 0.30 & 0.33 & 176 & 535 &  &  & - & 1,394 & \textbf{1,361} & \underline{1,365} & 1,590 & - & -12.33 & -14.45 & -14.18\\
        10 & 0.01 & 0.47 & 187 & 402 & 824 & 0.09 & - & 1,160 & \textbf{951} & \underline{980} & 1,525 & - & -23.95 & -37.65 & -35.75\\
        10 & 0.10 & 0.42 & 85 & 201 &  &  & - & 1,093 & \textbf{930} & \underline{950} & 1,370 & - & -20.25 & -32.11 & -30.67\\
        10 & 0.30 & 0.33 & 124 & 377 &  &  & - & 990 & \textbf{890} & \underline{901} & 1,285 & - & -22.98 & -30.73 & -29.92\\
        15 & 0.01 & 0.46 & 415 & 893 & 678 & 0.07 & - & 1,045 & \textbf{810} & \underline{841} & 1,488 & - & -29.76 & -45.54 & -43.45\\
        15 & 0.10 & 0.42 & 406 & 961 &  &  & - & 925 & \textbf{786} & \underline{813} & 1,379 & - & -32.96 & -43.01 & -41.07\\
        15 & 0.30 & 0.33 & 208 & 633 &  &  & - & 881 & \textbf{741} & \underline{759} & 1,147 & - & -23.23 & -35.38 & -33.86\\
        \midrule
        \multicolumn{5}{l}{Average} &  &  &  &  &  &  &  & - & -17.36 & -24.16 & -23.22\\
        \midrule
        \multicolumn{16}{l}{(-) no feasible solution; (\textbf{bold}) best result; (\underline{underline}) 2nd best result}\\
        \bottomrule
    \end{tabular}
\end{table}

\begin{table}[ht!]
    \caption{Running times for the \bankforty\ dataset.}
    \label{tab:bank_runtime_results}
    \footnotesize
    \setlength{\tabcolsep}{5.9pt}
    \begin{tabular}{lrrrrrrrrrrrrrr}
        \toprule
         &  &  &  &  & \multicolumn{5}{l}{$\text{CPU}_{\text{tot}}$ [s]} & \multicolumn{3}{l}{$\text{CPU}_{\text{avg}}$ [s]} & \multicolumn{2}{l}{$\text{Gap}_{\text{CPU}_{\text{avg}}}^{\text{SFC}}$ [\%]}\\
        \cmidrule(lr){6-10}\cmidrule(lr){11-13}\cmidrule(lr){14-15}
        $k$ & $\lambda$ & Target & $p$ & $q$ & MIQCP & SetVars & MPFC & Flow & SFC & MPFC & Flow & SFC & MPFC & Flow\\
        \midrule
        2 & 0.01 & 0.47 & 67 & 144 & limit & limit & 2,105.91 & 176.85 & limit & 21.06 & 1.77 & 37.41 & -43.70 & -95.27\\
        2 & 0.10 & 0.42 & 335 & 792 & limit & limit & 2,023.17 & 178.50 & limit & 20.23 & 1.79 & 147.55 & -86.29 & -98.79\\
        2 & 0.30 & 0.33 & 126 & 383 & limit & limit & 1,971.85 & 178.99 & limit & 19.72 & 1.79 & 58.09 & -66.05 & -96.92\\
        5 & 0.01 & 0.47 & 375 & 806 & limit & limit & limit & 306.56 & limit & 100.16 & 3.07 & 182.74 & -45.19 & -98.32\\
        5 & 0.10 & 0.42 & 291 & 688 & limit & limit & limit & 328.54 & limit & 85.86 & 3.29 & 131.41 & -34.66 & -97.50\\
        5 & 0.30 & 0.33 & 176 & 535 & limit & limit & limit & 408.30 & limit & 87.99 & 4.08 & 75.72 & 16.20 & -94.61\\
        10 & 0.01 & 0.47 & 187 & 402 & limit & limit & limit & 367.43 & limit & 362.67 & 3.67 & 95.78 & 278.65 & -96.17\\
        10 & 0.10 & 0.42 & 85 & 201 & limit & limit & limit & 454.45 & limit & 278.47 & 4.54 & 44.69 & 523.11 & -89.84\\
        10 & 0.30 & 0.33 & 124 & 377 & limit & limit & limit & 600.92 & limit & 278.06 & 6.01 & 57.32 & 385.10 & -89.52\\
        15 & 0.01 & 0.46 & 415 & 893 & limit & limit & limit & 389.36 & limit & 452.15 & 3.89 & 201.00 & 124.95 & -98.06\\
        15 & 0.10 & 0.42 & 406 & 961 & limit & limit & limit & 431.87 & limit & 361.42 & 4.32 & 174.17 & 107.51 & -97.52\\
        15 & 0.30 & 0.33 & 208 & 633 & limit & limit & limit & 593.17 & limit & 330.00 & 5.93 & 86.85 & 279.97 & -93.17\\
        \midrule
        \multicolumn{3}{l}{Average} & & &  &  &  &  &  &  &  &  & 119.97 & -95.47\\
        \midrule
        \multicolumn{15}{l}{(limit) time limit of 3{,}600 seconds reached}\\
        \bottomrule
    \end{tabular}
\end{table}

Table~\ref{tab:bank_cost_results} reports the results regarding solution quality for the instances of the \bankforty\ dataset. Each row of the table corresponds to one instance. The first two columns state the number of clusters ($k$) and the tolerance parameter ($\lambda$). The third column states the target balance (Target) resulting from the parameter $\lambda$. Columns four and five state the values of parameters $p$ and $q$, which are used by the benchmark approach SFC to control the target level of fairness. These parameters are obtained using Algorithm~\ref{algo:target_pq} in \ref{app:algorithm} based on the target balance (Target). Note that the target balance values in column three are rounded to two decimal places. As a result, some values appear identical even though their unrounded values differ. This explains why the table shows different values of $p$ and $q$ for the same (rounded) target balance. Columns six and seven report the clustering cost and the clustering balance achieved with Lloyd, respectively. Because Lloyd does not consider fairness, it generates the same solutions for different values of $\lambda$. We therefore report the results only for $\lambda=0.01$. Columns eight to twelve report the clustering costs obtained with MIQCP, SetVars, MPFC, MS-FlowFC (abbreviated as “Flow” in the table), and SFC. For each instance, we state the lowest clustering costs in bold and underline the second-lowest clustering costs. A dash (-) indicates that the respective approach could not find a feasible solution within the time limit of one hour. Columns thirteen to sixteen report the relative gaps of the clustering costs obtained with MIQCP, SetVars, MPFC, and MS-FlowFC with respect to the clustering costs obtained with the benchmark method SFC. A negative value indicates that the respective method achieved a lower clustering cost compared to SFC. The clustering balance of solutions obtained using fair clustering methods is not reported, as it consistently matches the target balance. As noted in Section~\ref{subsec:experimental_design}, the methods Lloyd, MPFC, MS-FlowFC, and SFC were each run up to 100 times with different random seeds. The reported clustering cost is the best value obtained across runs, and the reported running time is the total time over all runs performed. We can draw the following main conclusions from Table~\ref{tab:bank_cost_results}: 

\begin{itemize}[noitemsep, topsep=0pt]
    \item MPFC achieves the best overall solution quality, closely followed by MS-FlowFC. Both methods consistently outperform all other methods across all instances, in some cases by a substantial margin. Moreover, the relative advantage of the proposed methods tends to grow with the number of clusters ($k$).   
    \item Notably, the third-best method in terms of solution quality is SetVars. For each instance, it outperforms the benchmark method SFC. MIQCP is never able to find a feasible solution within the time limit.
    \item The clustering costs achieved by MPFC and MS-FlowFC are close to those obtained with Lloyd’s algorithm, particularly for small numbers of clusters ($k$). This further highlights the high solution quality of both methods. As the number of clusters increases, the clustering cost gap tends to widen. However, the solutions produced by Lloyd’s algorithm become increasingly unfair. 
\end{itemize}

Table~\ref{tab:bank_runtime_results} reports the results regarding running time for the instances of the \bankforty\ dataset. Each row of the table corresponds to one instance. The first five columns of Table~\ref{tab:bank_runtime_results} provide the same information about the instances as the first five columns of Table~\ref{tab:bank_cost_results}. Columns six to ten report the total running time of MIQCP, SetVars, MPFC, MS-FlowFC (abbreviated as “Flow” in the table), and SFC. The entry (limit) indicates that the time limit of one hour was reached. Columns eleven to thirteen report, for each approach that performs multiple runs, the average running time of a single run. Columns fourteen and fifteen report the relative gaps in terms of average running time of a single run of MPFC and MS-FlowFC compared to SFC. A negative value indicates that, on average, a single run of the respective approach is faster than a single run of SFC. We can draw the following main conclusions from Table~\ref{tab:bank_runtime_results}: 

\begin{itemize}[noitemsep, topsep=0pt]
    \item MS-FlowFC is by far the fastest algorithm, solving all instances well within the time limit. The running time of MS-FlowFC for a single run is, on average, 95.47\% lower than that of the benchmark algorithm SFC. 
    \item MPFC is the second-fastest algorithm for small numbers of clusters (small values of the parameter $k$), but its running time increases with increasing values of $k$. For $k\geq 10$, its running time per run exceeds, on average, that of SFC. 
    \item The running time of SFC appears to be affected by the parameters $p$ and $q$. The larger the values of $p$ and $q$, the higher the running time of SFC. 
    \item The exact approaches MIQCP and SetVars always reach the time limit. 
\end{itemize}

\begin{figure}[t]
    \centering
    \includegraphics[width=1\textwidth]{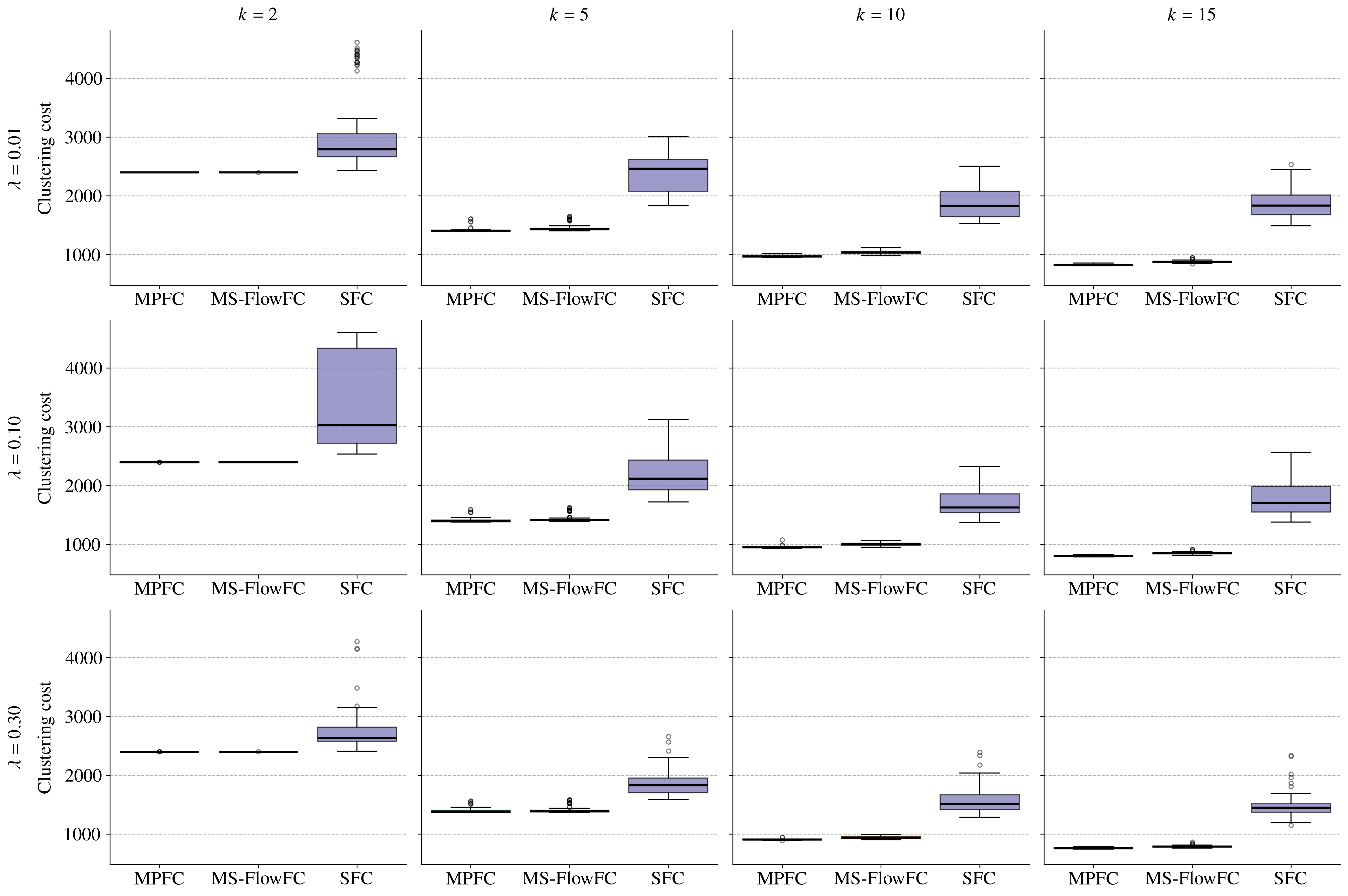}
    \caption{Boxplots of clustering costs for solutions obtained with different random seeds on the \bankforty\ dataset, across different methods, numbers of clusters, and tolerance values.}
    \label{fig:boxplots}
\end{figure}

In addition to solution quality and running time, we also analyze how the solution quality depends on the random seed. This analysis concerns only MPFC, MS-FlowFC, and SFC, as MIQCP and SetVars do not depend on a random seed. Fig.~\ref{fig:boxplots} shows twelve plots, one for each instance of the \bankforty\ dataset. The figure is organized so that all plots in the same row share the same value of the tolerance parameter ($\lambda$), while all plots in the same column share the same number of clusters ($k$). Each plot contains three boxplots, one for each method. The figure is based on the results reported in Table~\ref{tab:bank_cost_results} and Table~\ref{tab:bank_runtime_results}, i.e., a time limit of 3{,}600 seconds was used and each method was applied up to 100 times to each instance. We can draw the following main conclusions from Fig.~\ref{fig:boxplots}:

\begin{itemize}[noitemsep, topsep=0pt]
    \item The quality of solutions produced by SFC varies considerably with the random seed. Therefore, running SFC multiple times appears to be important to avoid obtaining low-quality solutions. 
    \item MPFC and MS-FlowFC show substantially lower variability in solution quality across random seeds than SFC. While multiple runs can further improve their solution quality, a single run with a fixed seed typically already yields a solution that has lower clustering cost than the best SFC solution obtained over multiple runs. 
    \item MPFC and MS-FlowFC usually require significantly fewer runs to match or surpass the solution quality of SFC, potentially resulting in shorter running times. 
\end{itemize}

\subsection{Results for large-scale instances}\label{subsec:large_results}

The results for the large-scale instances are consistent across the two datasets \swat\ and \census. We present here only the results for the instances derived from the \census\ dataset. The complete results are available on GitHub$^{\ref{github:results}}$. 

Table~\ref{tab:census_cost_results} reports the results regarding solution quality for the instances of the \census\ dataset. The table reports results only for the methods S-MPFC (the scalable version of MPFC), MS-FlowFC, SFC, and Lloyd, as the exact methods MIQCP and SetVars, as well as MPFC, are not suitable for such large-scale instances. The table follows the same structure as Table~\ref{tab:bank_cost_results}, described in the previous section. We can draw the following main conclusions from Table~\ref{tab:census_cost_results}: 

\begin{itemize}[noitemsep, topsep=0pt]
    \item MS-FlowFC and S-MPFC both outperform the benchmark method SFC in terms of solution quality, with MS-FlowFC performing best overall.
    \item Lloyd achieves the target balance only for $k=2$, and notably, MS-FlowFC yields a clustering cost nearly identical to that of Lloyd in this case. In all other instances, Lloyd fails to reach the target balance, whereas the fair clustering methods consistently satisfy it.
\end{itemize}

Table~\ref{tab:census_runtime_results} reports the results regarding running time for the instances of the \census\ dataset. The table follows the same structure as Table~\ref{tab:bank_runtime_results}, described in the previous section. The average running time of a single run of S-MPFC, reported in column eight, is computed as follows. We first determine the average running time across all 100 runs excluding the preprocessing time and then add the preprocessing time to the average running time (although it is performed only once at the beginning as described in Section~\ref{subsec:experimental_design}). This allows a fair comparison to the average running time reported for SFC, where the fairlet decomposition is performed in each run. We can draw the following main conclusions from Table~\ref{tab:census_runtime_results}: 

\begin{itemize}[noitemsep, topsep=0pt]
    \item S-MPFC is clearly the fastest fair clustering method for such large-scale instances. It consistently completes the 100 runs within a few minutes, whereas the other methods (MS-FlowFC and SFC) always reach the time limit of 3{,}600 seconds. Regarding the average running time per run, S-MPFC is, on average, $99.70\%$ faster than SFC.   
    \item The average running time per run of S-MPFC shows little sensitivity to the number of clusters ($k$) and the target balance (Target). In contrast, the running time of SFC appears to increase with increasing values of $p$ and $q$, while that of MS-FlowFC appears to grow with $k$. 
\end{itemize}

In the results for the \swat\ dataset, available on GitHub$^{\ref{github:results}}$, we observed that running S-MPFC using only $r=100$ representatives can sometimes lead to increased running times. This occurred for the two instances with $k \in \{10,15\}$ and $\text{Target}=0.06$. The \swat\ dataset is special because the two protected groups of the sensitive feature \texttt{Normal/Attack} are highly imbalanced. Only a small number of objects belong to the protected group ``Attack’’, while the vast majority belong to the protected group ``Normal’’. This imbalance results in a low dataset balance of 0.061. When only $r=100$ representatives are constructed, only a few representatives may represent objects from the protected group ``Attack’’. Therefore, finding a feasible clustering becomes challenging when parameter $k$ is relatively large and the target balance is very close to the dataset balance, as in the case of $k \in \{10,15\}$ and $\text{Target}=0.06$. Consequently, in some runs, the Gurobi Optimizer may spend considerable time searching for a feasible solution, which explains the two outliers in running time. This issue could be mitigated by increasing the number of representatives~$r$.

\begin{table}[t!]
    \caption{Clustering quality results for the \census\ dataset.}
    \label{tab:census_cost_results}
    \footnotesize
    \setlength{\tabcolsep}{10.2pt}
    \begin{tabular}{lrrrrrrrrrr}
        \toprule
        & & & & \multicolumn{2}{l}{Lloyd} & \multicolumn{3}{l}{Cost} & \multicolumn{2}{l}{$\text{Gap}_{\text{Cost}}^{\text{SFC}}$ [\%]} \\
        \cmidrule(lr){5-6}\cmidrule(lr){7-9}\cmidrule(lr){10-11}
        $k$ & Target & $p$ & $q$ & Cost & Balance & S-MPFC & Flow & SFC & S-MPFC & Flow\\
        \midrule
        2 & 0.90 & 9 & 10 & 3,205,963 & 0.93 & \underline{3,234,760} & \textbf{3,206,079} & 3,324,154 & -2.69 & -3.55\\
        2 & 0.75 & 3 & 4 &  &  & \underline{3,234,760} & \textbf{3,206,079} & 3,236,939 & -0.07 & -0.95\\
        2 & 0.50 & 1 & 2 &  &  & 3,234,760 & \textbf{3,206,080} & \underline{3,213,830} & 0.65 & -0.24\\
        5 & 0.90 & 9 & 10 & 2,082,991 & 0.49 & \underline{2,436,024} & \textbf{2,229,905} & 2,594,286 & -6.10 & -14.05\\
        5 & 0.75 & 3 & 4 &  &  & \underline{2,267,625} & \textbf{2,142,045} & 2,464,672 & -7.99 & -13.09\\
        5 & 0.50 & 1 & 2 &  &  & \underline{2,122,421} & \textbf{2,083,342} & 2,388,713 & -11.15 & -12.78\\
        10 & 0.90 & 9 & 10 & 1,619,760 & 0.46 & \underline{2,038,966} & \textbf{1,801,850} & 2,218,436 & -8.09 & -18.78\\
        10 & 0.75 & 3 & 4 &  &  & \underline{1,882,197} & \textbf{1,712,834} & 2,055,310 & -8.42 & -16.66\\
        10 & 0.50 & 1 & 2 &  &  & \underline{1,717,188} & \textbf{1,639,519} & 1,900,991 & -9.67 & -13.75\\
        15 & 0.90 & 9 & 10 & 1,394,466 & 0.36 & \underline{1,902,220} & \textbf{1,613,071} & 2,027,372 & -6.17 & -20.44\\
        15 & 0.75 & 3 & 4 &  &  & \underline{1,740,733} & \textbf{1,524,018} & 1,885,811 & -7.69 & -19.19\\
        15 & 0.50 & 1 & 2 &  &  & \underline{1,523,456} & \textbf{1,435,454} & 1,758,309 & -13.36 & -18.36\\
        \midrule
        \multicolumn{3}{l}{Average} &  &  &  &  &  &  & -6.73 & -12.65\\
        \midrule
        \multicolumn{11}{l}{(\textbf{bold}) best result; \ (\underline{underline}) 2nd best result}\\
        \bottomrule
    \end{tabular}
\end{table}

\begin{table}[ht!]
    \caption{Running times for the \census\ dataset.}
    \label{tab:census_runtime_results}
    \footnotesize
    \setlength{\tabcolsep}{10.2pt}
    \begin{tabular}{lrrrrrrrrrrr}
        \toprule
         &  &  &  & \multicolumn{3}{l}{$\text{CPU}_{\text{tot}}$ [s]} & \multicolumn{3}{l}{$\text{CPU}_{\text{avg}}$ [s]} & \multicolumn{2}{l}{$\text{Gap}_{\text{CPU}_{\text{avg}}}^{\text{SFC}}$ [\%]}\\
        \cmidrule(lr){5-7}\cmidrule(lr){8-10}\cmidrule(lr){11-12}
        $k$ & Target & $p$ & $q$ & S-MPFC & Flow & SFC & $\text{S-MPFC}^\dagger$ & Flow & SFC & S-MPFC & Flow\\
        \midrule
        2 & 0.90 & 9 & 10 & 14.85 & limit & limit & 1.89 & 90.08 & 851.35 & -99.78 & -89.42\\
        2 & 0.75 & 3 & 4 & 14.55 & limit & limit & 1.88 & 84.13 & 739.70 & -99.75 & -88.63\\
        2 & 0.50 & 1 & 2 & 16.24 & limit & limit & 1.90 & 84.08 & 751.03 & -99.75 & -88.80\\
        5 & 0.90 & 9 & 10 & 44.92 & limit & limit & 2.19 & 180.14 & 856.94 & -99.74 & -78.98\\
        5 & 0.75 & 3 & 4 & 38.63 & limit & limit & 2.13 & 200.31 & 746.86 & -99.71 & -73.18\\
        5 & 0.50 & 1 & 2 & 30.40 & limit & limit & 2.04 & 302.46 & 721.01 & -99.72 & -58.05\\
        10 & 0.90 & 9 & 10 & 107.60 & limit & limit & 2.82 & 458.22 & 850.03 & -99.67 & -46.09\\
        10 & 0.75 & 3 & 4 & 68.24 & limit & limit & 2.42 & 450.19 & 745.43 & -99.68 & -39.61\\
        10 & 0.50 & 1 & 2 & 51.96 & limit & limit & 2.26 & 453.31 & 721.92 & -99.69 & -37.21\\
        15 & 0.90 & 9 & 10 & 166.27 & limit & limit & 3.40 & 522.20 & 846.87 & -99.60 & -38.34\\
        15 & 0.75 & 3 & 4 & 102.86 & limit & limit & 2.77 & 528.96 & 737.56 & -99.62 & -28.28\\
        15 & 0.50 & 1 & 2 & 66.71 & limit & limit & 2.41 & 739.10 & 719.03 & -99.66 & 2.79\\
        \midrule
        \multicolumn{3}{l}{Average} & &  &  &  &  &  &  & -99.70 & -55.32\\
        \midrule
        \multicolumn{12}{l}{$^\dagger$ including preprocessing time; \ (limit) time limit of 3{,}600 seconds reached}\\
        \bottomrule
    \end{tabular}
\end{table}

\subsection{Cost-fairness trade-off experimental results}\label{subsec:tradeoff_results}

In this section, we present the results of the cost-fairness trade-off analysis, in which we compare the performance of our heuristics MPFC and MS-FlowFC with the benchmark approach O\&C for different desired levels of fairness. As described in Section~\ref{subsec:experimental_design}, we first applied the benchmark approach O\&C to the \bankfive\ dataset using four different values of the number of clusters ($k$) and four different values of the desired level of fairness ($\alpha$), resulting in sixteen instances. The clustering balance of the resulting solutions was used as the target balance for our heuristics MPFC and MS-FlowFC. Note that for this analysis, we increased the time limit from 3{,}600 to 10{,}800 seconds.

Table~\ref{tab:tradeoff_results} reports the results of the comparison between O\&C, MPFC, and MS-FlowFC. Each row of the table corresponds to one instance. The first two columns state the number of clusters ($k$) and the desired fairness level ($\alpha$). Columns three to five report the clustering cost obtained with O\&C, MPFC, and MS-FlowFC (abbreviated as “Flow” in the table). For each instance, we state the lowest clustering costs in bold and underline the second-lowest clustering costs. Columns six and seven report the relative gap of the clustering costs obtained with MPFC and MS-FlowFC with respect to the clustering costs obtained with the benchmark method O\&C. A negative value indicates that the respective method achieved a lower clustering cost compared to O\&C. Column eight reports the clustering balance of the solutions obtained with O\&C. Columns nine and ten indicate, with a check mark, that MPFC and MS-FlowFC obtained solutions with a clustering balance equal to or higher than that of the solutions obtained with O\&C. Columns eleven to thirteen report the total running time for each method. As described in Section~\ref{subsec:experimental_design}, the methods MPFC and MS-FlowFC were each run up to 100 times with different random seeds. The reported clustering cost is the best value obtained across runs, and the reported running time is the total time over all runs performed. The method O\&C does not rely on a random seed and was therefore applied only once. We can draw the following main conclusions from Table~\ref{tab:tradeoff_results}:

\begin{itemize}[noitemsep, topsep=0pt]
    \item MPFC and MS-FlowFC produce solutions with equal or higher clustering balance than those produced by the benchmark method O\&C, while having substantially lower clustering costs. On average, both MPFC and MS-FlowFC reduce the clustering costs of O\&C by 24.33\%.
    \item When the desired fairness level $\alpha = 0$, which corresponds to vanilla clustering, the clustering costs obtained by the three methods are the same or very close. However, once the desired fairness level increases ($\alpha > 0$), O\&C is outperformed by the proposed approaches, with the cost gap widening significantly as $k$ increases.
    \item MS-FlowFC is by far the fastest method, completing 100 runs in a few seconds to a few minutes, whereas the benchmark method O\&C always requires more than two hours per instance. MPFC is the second-fastest method, consistently completing 100 runs in under two hours. 
\end{itemize}

\begin{table}[t]
    \caption{Results of the comparison between O\&C, MPFC, and MS-FlowFC on the \bankfive\ dataset.}
    \label{tab:tradeoff_results}
    \footnotesize
    \setlength{\tabcolsep}{8.2pt}
    \begin{tabular}{lrrrrrrrrrrrr}
        \toprule
        & & \multicolumn{3}{l}{Cost} & \multicolumn{2}{l}{$\text{Gap}_{\text{Cost}}^{\text{O\&C}}$ [\%]} & \multicolumn{3}{l}{Balance} & \multicolumn{3}{l}{CPU [s]} \\
        \cmidrule(lr){3-5}\cmidrule(lr){6-7}\cmidrule(lr){8-10}\cmidrule(lr){11-13}
        $k$ & $\alpha$ & O\&C & MPFC & Flow & MPFC & Flow & O\&C & MPFC & Flow & O\&C & MPFC & Flow\\
        \midrule
        2 & 0.0 & \textbf{325.43} & \textbf{325.43} & \textbf{325.43} & 0.00 & 0.00 & 0.183 & \checkmark & \checkmark & 10,087.1 & 466.7 & 43.8\\
        2 & 0.5 & \textbf{325.65} & \textbf{325.65} & \underline{325.68} & 0.00 & 0.01 & 0.191 & \checkmark & \checkmark & 9,435.8 & 434.5 & 40.6\\
        2 & 1.5 & \underline{332.86} & \textbf{325.65} & \textbf{325.65} & -2.17 & -2.17 & 0.190 & \checkmark & \checkmark & 9,377.8 & 434.4 & 40.7\\
        2 & 2.0 & \underline{335.44} & \textbf{325.65} & \textbf{325.65} & -2.92 & -2.92 & 0.190 & \checkmark & \checkmark & 9,773.5 & 434.4 & 40.7\\
        5 & 0.0 & \textbf{182.82} & \underline{182.87} & 182.89 & 0.03 & 0.04 & 0.120 & \checkmark & \checkmark & 10,220.9 & 1,714.6 & 125.5\\
        5 & 0.5 & 208.90 & \textbf{183.60} & \underline{183.67} & -12.11 & -12.08 & 0.190 & \checkmark & \checkmark & 9,658.0 & 1,841.5 & 121.8\\
        5 & 1.5 & \underline{347.17} & \textbf{183.56} & \textbf{183.56} & -47.13 & -47.13 & 0.189 & \checkmark & \checkmark & 9,349.0 & 1,817.1 & 126.4\\
        5 & 2.0 & 336.21 & \textbf{183.40} & \underline{183.45} & -45.45 & -45.44 & 0.183 & \checkmark & \checkmark & 9,651.2 & 1,769.8 & 124.4\\
        10 & 0.0 & \textbf{115.91} & \underline{115.97} & \underline{115.97} & 0.05 & 0.06 & 0.068 & \checkmark & \checkmark & 10,249.1 & 4,132.8 & 261.8\\
        10 & 0.5 & 203.31 & \underline{116.63} & \textbf{116.62} & -42.63 & -42.64 & 0.161 & \checkmark & \checkmark & 9,984.1 & 4,399.4 & 241.9\\
        10 & 1.5 & 215.79 & \underline{116.24} & \textbf{116.22} & -46.13 & -46.14 & 0.140 & \checkmark & \checkmark & 9,543.2 & 4,340.7 & 251.9\\
        10 & 2.0 & 218.00 & \underline{116.09} & \textbf{116.06} & -46.75 & -46.76 & 0.118 & \checkmark & \checkmark & 9,861.9 & 4,252.1 & 250.3\\
        15 & 0.0 & 97.32 & \textbf{96.85} & \underline{97.07} & -0.49 & -0.25 & 0.017 & \checkmark & \checkmark & 10,058.6 & 6,478.4 & 293.1\\
        15 & 0.5 & 163.36 & \underline{96.85} & \textbf{96.72} & -40.72 & -40.80 & 0.000 & \checkmark & \checkmark & 9,489.0 & 6,397.4 & 288.8\\
        15 & 1.5 & 198.24 & \underline{96.85} & \textbf{96.72} & -51.15 & -51.21 & 0.000 & \checkmark & \checkmark & 9,877.5 & 6,397.4 & 288.8\\
        15 & 2.0 & 202.39 & \underline{97.58} & \textbf{97.53} & -51.79 & -51.81 & 0.150 & \checkmark & \checkmark & 9,710.4 & 7,022.5 & 279.6\\
        \midrule
        \multicolumn{3}{l}{Average} &  &  & -24.33 & -24.33 &  &  &  &  &  & \\
        \midrule
        \multicolumn{13}{l}{(\textbf{bold}) best result; \ (\underline{underline}) 2nd best result; \ (\checkmark) O\&C's clustering balance attained}\\
        \bottomrule
    \end{tabular}
\end{table}

\begin{figure}[t]
    \centering
    \includegraphics[width=1\textwidth]{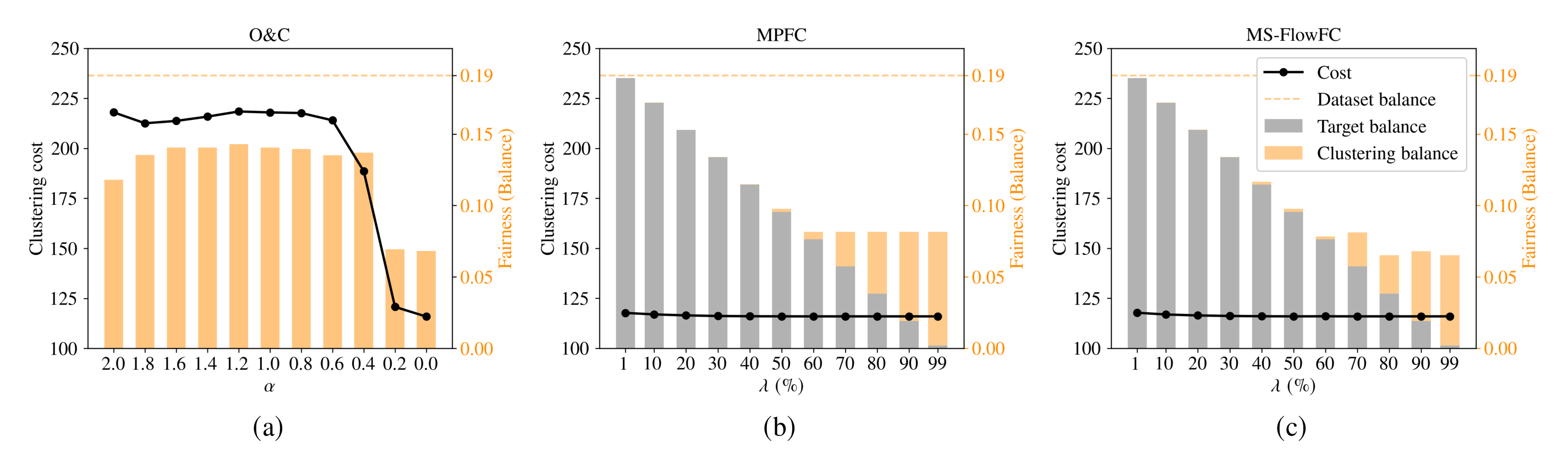}
    \caption{Cost-fairness trade-off of O\&C (a), MPFC (b), and MS-FlowFC (c) on the \bankfive\ dataset with $k=10$.}
    \label{fig:tradeoff}
\end{figure}

In Table~\ref{tab:tradeoff_results}, we can also observe that the choice of the desired fairness level $\alpha$ does not consistently affect the clustering balance. Hence, we conduct an additional analysis to evaluate in more detail how the respective fairness parameter controls the clustering balance. This analysis also allows us to study how effectively the different methods control the trade-off between clustering balance and clustering cost. Fig.~\ref{fig:tradeoff} presents the results of this analysis with three subplots, one for each method. Each subplot shows the clustering costs (black line) and the clustering balances (orange bars) for eleven different values of the fairness control parameter ($\alpha$ for O\&C and $\lambda$ for MPFC and MS-FlowFC). Note that the values of the control parameters ($\alpha$ and $\lambda$) are arranged along the horizontal axis of the subplots such that the level of fairness decreases from left to right. We can draw the following main conclusions from Fig.~\ref{fig:tradeoff}:

\begin{itemize}[noitemsep, topsep=0pt]
    \item MPFC and MS-FlowFC demonstrate precise control over the clustering balance via the tolerance parameter $\lambda$. For $\lambda = 0.01$ (1\% in the figure), both methods achieve clustering balances close to the maximum possible value (i.e., the dataset balance). As $\lambda$ increases, the clustering balances decrease approximately proportionally, eventually reaching the clustering balances of vanilla clustering solutions. In the plot, this occurs at $\lambda = 0.60$ (60\% in the figure) for MPFC, whereas MS-FlowFC shows slightly erratic clustering balance values.
    \item O\&C does not appear to offer precise control over the clustering balance via the fairness parameter $\alpha$. For $\alpha \in [0,1]$, O\&C yields varying levels of fairness, but these are not precisely controlled. For example, the improvement in balance from $\alpha = 0.2$ to $\alpha = 0.4$ is considerably larger than that from $\alpha = 0$ to $\alpha = 0.2$ or from $\alpha = 0.4$ to $\alpha = 0.6$. For $\alpha > 1$, O\&C is unable to control the clustering balance.
    \item The clustering cost curves of MPFC and MS-FlowFC are nearly flat, demonstrating that these methods are able to achieve fair solutions with only a marginal increase in clustering cost. Even when the clustering balance is nearly equal to the dataset balance ($\lambda=0.01$), the cost of solutions obtained with MPFC and MS-FlowFC remains substantially lower than that of solutions obtained with O\&C.
\end{itemize}

\section{Conclusions}\label{sec:conclusions}

In this paper, we introduced three new heuristics (MPFC, MS-FlowFC, and S-MPFC) for fair $k$-means clustering. All three heuristics employ the $k$-means decomposition scheme, in which an object assignment step and a cluster center update step are performed iteratively until convergence. The methods mainly differ in how the assignment step is implemented. MPFC solves a binary linear program. MS-FlowFC solves multiple minimum-cost flow problems. S-MPFC employs a preprocessing technique that enables solving an aggregated version of the binary linear program used in MPFC. Each of the proposed heuristics offers advantages over existing methods. MPFC is flexible enough to accommodate application-specific requirements. MS-FlowFC is more scalable than MPFC and does not rely on an integer programming solver. S-MPFC is highly scalable thanks to its preprocessing technique. Moreover, all heuristics offer effective control over the cost-fairness trade-off. In addition, we proposed two exact methods to benchmark the performance of the heuristics. The results of numerical experiments demonstrated that the proposed heuristics can consistently outperform leading fair clustering methods in terms of solution quality, running time, and control over the cost-fairness trade-off. 

There are several promising directions for future research. First, the proposed heuristics could be extended to incorporate additional constraints. In MPFC and S-MPFC, such constraints could be added directly to the binary linear programming model. In MS-FlowFC, the flow network used in the assignment step could be adapted to consider additional constraints such as, for example, cardinality constraints on the clusters. Second, future work could investigate alternative strategies to determine the initial positions of the cluster centers. For example, it may be beneficial to consider information about the protected groups when selecting the initial cluster positions, or to ensure that the initial positions of different runs are sufficiently distinct. Finally, the techniques presented in this paper could be applied to related problems, such as the fair facility location problem.

    \appendix
    \section{Exact approaches}\label{app:exact}

\subsection{Mixed-integer quadratically constrained programming model (MIQCP)}\label{subsec:miqcp}

The first exact method is based on the mixed-integer quadratically constrained programming model presented in formulation (MIQCP). In this formulation, $c_{ij}$ is a decision variable representing the distance between object $i$ and center $j$, while $\mu_{jf}$ is a decision variable representing the $f$-th coordinate of cluster center $j$. Each parameter $v_{if}$ denotes the value of the $f$-th non-sensitive feature of object $i$, while $M$ is a sufficiently large number. The objective function in (\ref{miqcp_a}) minimizes the $k$-means clustering cost. Constraints (\ref{miqcp_b}) ensure that the distance between object $i$ and center $j$ is considered only when the corresponding assignment variable is equal to one. Constraints (\ref{miqcp_c}) guarantee that each object is assigned to exactly one cluster, while constraints (\ref{miqcp_d}) ensure that no cluster remains empty. Constraints (\ref{miqcp_e}) enforce fairness by imposing a lower bound on the balance of each cluster with respect to any sensitive feature. Finally, constraints (\ref{miqcp_f}), (\ref{miqcp_g}), and (\ref{miqcp_h}) define the domain of the decision variables.
\begin{empheq}[left=\text{(MIQCP)}\empheqlbrace]{alignat=2}
  \min\text{} & \sum_{i=1}^n \sum_{j=1}^k c_{ij} & & \label{miqcp_a}\\
  \text{s.t. } & c_{ij} \geq \sum_{f=1}^d (v_{if} - \mu_{jf})^2 - M(1 - x_{ij}) & \quad & (i=1,\ldots,n;\ j=1,\ldots,k) \label{miqcp_b}\\
  & \sum_{j=1}^k x_{ij} = 1 & \quad & (i=1,\ldots,n) \label{miqcp_c}\\
  & \sum_{i=1}^n x_{ij} \geq 1 & \quad & (j=1,\ldots,k) \label{miqcp_d}\\
  & \sum_{i \in G_{gs}} x_{ij} \geq \overline{B}_s(X, \lambda) \sum_{i \in G_{g^{\prime}s}} x_{ij} & \quad & (g,g' \in \mathcal{G}_s: g \ne g';\ j=1,\ldots,k;\ s \in S) \label{miqcp_e}\\
  & x_{ij} \in \{0, 1\} & \quad & (i=1,\ldots,n;\ j=1,\ldots,k) \label{miqcp_f}\\
  & c_{ij} \geq 0 & \quad & (i=1,\ldots,n;\ j=1,\ldots,k) \label{miqcp_g}\\
  & \mu_{jf} \in \mathbb{R} & \quad & (j=1,\ldots,k;\ f=1,\ldots,d) \label{miqcp_h}
\end{empheq}

\subsection{Set-variable-based model (SetVars)}\label{subsec:setvars}

The second exact method, SetVars, is formulated using the set-variable-based modeling framework available in the commercial solver Hexaly. A formulation for the vanilla $k$-means clustering problem is provided in \cite{hexalykmeans} and is extended here by incorporating fairness constraints, as shown in formulation (SetVars). Unlike MIQCP, which uses binary variables to assign objects to clusters, SetVars uses a set variable $C_j$ for each cluster $j$. The $f$-th coordinate of cluster center $j$ is denoted by $\mu_{jf}$, while the value of the $f$-th non-sensitive feature of object $i$ is denoted by $v_{if}$. The objective function in (\ref{setvars_a}) minimizes the $k$-means clustering cost, where the coordinates of cluster centers are defined in Eqs.~(\ref{setvars_g}). The set variables for clusters are defined in Eqs.~(\ref{setvars_b}). Constraint (\ref{setvars_c}) and constraints (\ref{setvars_d}) jointly ensure that all objects in the dataset are assigned to a cluster and that there are no overlaps between cluster sets. Constraints (\ref{setvars_e}) impose a non-emptiness condition on cluster sets. Finally, the fairness constraints (\ref{setvars_f}) impose a lower bound on the balance of each cluster with respect to any sensitive feature.

\begin{empheq}[left=\text{(SetVars)}\empheqlbrace]{alignat=2}
  \min\text{} & \sum_{j=1}^k \sum_{i \in C_j} \sum_{f=1}^d \left(v_{if} - \mu_{jf} \right)^2 \label{setvars_a}\\
  \text{s.t. } & C_j = \text{SetVar}(n) & \quad & (j=1,\ldots,k) \label{setvars_b}\\
  & \sum_{j=1}^{k} |C_j| = n & \quad & \label{setvars_c}\\
  & C_j \cap C_{j^{\prime}} = \emptyset & \quad & (j, j^{\prime} = 1,\ldots,k: j \ne j^{\prime}) \label{setvars_d}\\
  & |C_j| \geq 1 & \quad & (j=1,\ldots,k) \label{setvars_e}\\
  & |C_j \cap G_{gs}| \geq \overline{B}_s(X, \lambda)\, |C_j \cap G_{g^{\prime}s}| & \quad & (g,g' \in \mathcal{G}_s: g \ne g';\ j=1,\ldots,k;\ s \in S) \label{setvars_f}\\
  & \mu_{jf} = \frac{1}{|C_j|} \sum_{i \in C_j} v_{if} & \quad & (j=1,\ldots,k;\ f=1,\ldots,d) \label{setvars_g}
\end{empheq}

\section{Algorithm to derive parameters $p$ and $q$ for the $(p,q)$-fairlet decomposition}\label{app:algorithm}

Algorithm~\ref{algo:target_pq} computes the integer parameters $p$ and $q$ required for the $(p,q)$-fairlet decomposition of SFC, given the target balance $\overline{B}_s(X,\lambda)$ as input. The ratio $\frac{p}{q}$ serves as an approximation of the target balance $\overline{B}_s(X,\lambda)$. In Algorithm~\ref{algo:target_pq}, the parameter values used at each iteration are denoted by $\bar{p}$ and $\bar{q}$, while the final returned values are denoted by $p$ and $q$. The ratio obtained at each iteration is represented by $\bar{r}$, and the best ratio found so far is denoted by $r^{*}$. The algorithm initializes the best ratio by setting $\bar{p}=0$ and $\bar{q}=1$, yielding $r^{*}=\frac{0}{1}$. It then iterates over denominator values $\bar{q}=1,\dots,1000$. For each $\bar{q}$, the corresponding numerator is computed as $\bar{p}=\lfloor \bar{q} \cdot \overline{B}_s(X,\lambda)\rfloor$. The resulting ratio $\bar{r}=\frac{\bar{p}}{\bar{q}}$ is compared with the current best ratio $r^{*}$. Whenever $\bar{r}$ is greater than the current best ratio $r^{*}$, the algorithm updates $r^{*}$. Finally, Algorithm~\ref{algo:target_pq} returns the pair $(p,q)$ corresponding to the integer parameters defining the best ratio $r^{*}$ approximating the target balance.

\begin{algorithm}[h!]
\caption{Generation of the integers $p$ and $q$ for the $(p,q)$-fairlet decomposition}
\label{algo:target_pq}
    \begin{algorithmic}[1]
        \Procedure{GetFairletIntegers}{$\overline{B}_s(X, \lambda)$}
            \State best balance $r^* \gets \tfrac{0}{1}$;
            \For{denominator $\bar{q} = 1$ to $\bar{q}_{\max}=1000$}
                \State numerator $\bar{p} \gets \lfloor \bar{q} \cdot \overline{B}_s(X, \lambda) \rfloor$;
                \State candidate balance $\bar{r} \gets \tfrac{\bar{p}}{\bar{q}}$;
                \If{$\bar{r} > r^*$}
                    \State $r^* \gets \bar{r}$;
                \EndIf
            \EndFor
            \State $p \gets$ numerator of $r^*$;
            \State $q \gets$ denominator of $r^*$;
            \State \Return $(p, q)$
        \EndProcedure
    \end{algorithmic}
\end{algorithm}

    \bibliographystyle{elsarticle-harv} 
    \bibliography{bibliography}

\end{document}